\pdfoutput=1
\documentclass[11pt]{article}
\usepackage[table,dvipsnames]{xcolor}
\usepackage{times}
\usepackage{acl}
\usepackage{latexsym}
\usepackage[T1]{fontenc}
\usepackage[utf8]{inputenc}
\usepackage{microtype}
\usepackage{inconsolata}
\usepackage{amsmath}
\usepackage{amssymb}
\usepackage{graphicx}
\usepackage{enumerate}
\usepackage{balance}
\usepackage{blindtext}
\usepackage{listings}
\usepackage{booktabs}
\usepackage{colortbl}
\usepackage{tcolorbox}
\usepackage{marvosym}
\usepackage{float}
\usepackage{makecell}
\usepackage{multicol}
\usepackage{multirow}
\usepackage{tabularx}
\usepackage{marvosym}
\usepackage{stfloats}
\usepackage{graphicx}
\usepackage{subfigure}
\usepackage{subcaption}
\usepackage[normalem]{ulem}
\usepackage{svg}
\usepackage{hyperref}
\usepackage{csquotes}
\usepackage{epigraph}
\usepackage{wrapfig}
\usepackage{amsfonts}
\usepackage{bbm}
\usepackage{algorithm}

\usepackage{algorithmicx} 
\usepackage{algpseudocode} 
\usepackage{booktabs}
\usepackage{arydshln}
\usepackage{makecell}
\usepackage{etoc}

\tcbuselibrary{skins,breakable} 
\newcommand\footnoteONLYtext[1]{
    \let \mybackup \thefootnote
    \let \thefootnote \relax
    \footnotetext{#1}
    \let \thefootnote \mybackup
    \let \mybackup \imareallyundefinedcommand}

\lstdefinestyle{PythonCode}{
    language=Python,
    basicstyle=\ttfamily,
    breaklines=true,
    keywordstyle=\bfseries\color{NavyBlue},
    morekeywords={},
    emph={self},
    emphstyle=\bfseries\color{Rhodamine},
    commentstyle=\itshape\color{black!50!white},
    stringstyle=\bfseries\color{PineGreen!90!black},
    columns=flexible,
}

\lstdefinestyle{BashCode}{
    language=Bash,
    basicstyle=\ttfamily\color{white}, 
    backgroundcolor=\color{black},      
    breaklines=true,
    keywordstyle=\bfseries\color{MidnightBlue},
    morekeywords={},
    emph={},
    emphstyle=\bfseries\color{Purple},
    commentstyle=\itshape\color{black!50!white},
    stringstyle=\bfseries\color{OliveGreen!90!black},
    columns=flexible,
}

\newcommand{\ie}{\textit{i.e., }}
\newcommand{\eg}{\textit{e.g., }}

\newcommand{\iid}{\textit{i.i.d. }}

\definecolor{lightpurple}{RGB}{242, 242, 255}
\definecolor{lightyellow}{RGB}{252, 240, 209}
\definecolor{uncertainty}{RGB}{120,170,230}

\usepackage{xspace}
 
\newcommand{\eat}[1]{}

\newcommand{\modelname}{{Interlat}\xspace}

\newcommand{\agentbubble}[1]{
  \noindent
  \begin{minipage}[t]{0.75\linewidth}
    \begin{tcolorbox}[colback=blue!3!white,colframe=blue!50!black,boxrule=0.3pt,arc=1mm,enhanced]
      \textbf{Agent}\\[-2pt] #1
    \end{tcolorbox}
  \end{minipage}\par
}
\newcommand{\envbubble}[1]{
  \noindent\hfill
  \begin{minipage}[t]{0.75\linewidth}
    \begin{tcolorbox}[colback=gray!8!white,colframe=gray!60!black,boxrule=0.3pt,arc=1mm,enhanced]
      \textbf{Env}\\[-2pt] #1
    \end{tcolorbox}
  \end{minipage}\par
}

\newtcolorbox{templatebox}[1][]{
  enhanced,
  breakable,                   
  colback=black!3,              
  colframe=black,               
  boxrule=0.8pt,                
  arc=3mm,                      
  left=1em,right=1em,top=1em,bottom=0.7em,  
  overlay first={
    \node[fill=black, text=white, font=\small\bfseries, rounded corners=2pt,
          anchor=north west, inner sep=4pt, outer sep=0pt, yshift=1ex] 
          at ([xshift=12pt, yshift=0.5ex]frame.north west) {#1}; 
  },
  #1/.code={} 
}

\newtcolorbox{breakcard}[1][]{
  enhanced,
  breakable,                
  colback=cardbg,
  colframe=cardframe,
  boxrule=1pt,
  arc=4pt, outer arc=4pt,
  left=3mm, right=3mm, top=2mm, bottom=2mm,
  fonttitle=\bfseries,
  colbacktitle=cardtitle,
  coltitle=white,
  boxed title style={colframe=cardtitle, colback=cardtitle, boxrule=0pt, arc=4pt, outer arc=4pt},
  width=\linewidth,        
  before skip=8pt, after skip=8pt,
  #1
}

\definecolor{TrainRed}{HTML}{E74C3C}         
\definecolor{UntrainGray}{HTML}{7F7F7F}       
\definecolor{TrainRedBg}{HTML}{E57373}        
\definecolor{UntrainGrayBg}{HTML}{D3D3D3}     
\definecolor{UntrainBlueBg}{HTML}{005FAF}

\newcommand{\code}[1]{\texttt{#1}}

\newcommand{\imend}{\code{<im\_end>}}
\newcommand{\sys}{\code{<im\_start>system}}
\newcommand{\user}{\code{<im\_start>user}}
\newcommand{\asst}{\code{<im\_start>assistant}}

\author{\textbf{Zhuoyun Du}$^{1,2,3\dagger\text{*}}$ \quad \textbf{Runze Wang}$^{2\dagger}$ \quad \textbf{Huiyu Bai}$^{2}$ \quad \textbf{Zouying Cao}$^{4}$ \\ 
\textbf{Xiaoyong Zhu}$^{2}$ \quad \textbf{Yu Cheng}$^{2}$ \quad \textbf{Bo Zheng}$^{2}$ \quad \textbf{Wei Chen}$^{1}$ \quad \textbf{Haochao Ying}$^{5\text{\Letter}}$\\[3pt]
$^{1}$State Key Lab of CAD\&CG, Zhejiang University \\ 
$^{2}$Future Living Lab of Alibaba \\ $^{3}$Zhejiang Key Laboratory of Medical Imaging Artificial Intelligence\\ $^{4}$Shanghai Jiao Tong University \quad $^{5}$Zhejiang University \\[3pt]
\texttt{\{duzy, haochaoying\}@zju.edu.cn, yunze.wrz@alibaba-inc.com}
}

\title{Enabling Agents to Communicate Entirely in Latent Space}

\begin{document}

\maketitle
\footnoteONLYtext{$^\dagger$Equal Contribution.}
\footnoteONLYtext{$^*$Work done during an internship at Alibaba Group.}
\footnoteONLYtext{$^{\text{\Letter}}$Corresponding Authors.}

\begin{abstract}
While natural language is the de facto communication medium for LLM-based agents, it presents a fundamental constraint. The process of downsampling rich, internal latent states into discrete tokens inherently limits the depth and nuance of information that can be transmitted, thereby hindering collaborative problem-solving. 
Inspired by telepathy, which bypasses symbolic language in communication, we propose \textbf{\modelname} (\textit{\textbf{\underline{Inter}}-agent \textbf{\underline{Lat}}ent Space Communication}), a paradigm that leverages the continuous last hidden states of an LLM as a representation of its thought for direct communication (termed \textbf{\enquote{latent communication}}). An additional learned compression process further compresses latent communication via latent space reasoning. Experiments demonstrate that \modelname outperforms both fine-tuned chain-of-thought (CoT) prompting and single-agent baselines, even across heterogeneous models, promoting more exploratory behavior and enabling genuine utilization of latent information. Further compression not only substantially accelerates inference by \textbf{up to 24$\times$} but also maintains competitive performance through an efficient information-preserving mechanism. We position this work as a feasibility study of entirely latent space inter-agent communication, and our results highlight its potential, offering valuable insights for future research. 
Our code is available at~\href{https://github.com/XiaoDu-flying/Interlat}{https://github.com/XiaoDu-flying/Interlat}.
\end{abstract}

\begin{quote}
  \raggedleft
  \emph{\enquote{The limits of my language mean the limits of my world.}}\\
  --- Ludwig Wittgenstein,
  \emph{Tractatus Logico-Philosophicus}, \S 5.6.
\end{quote}

\section{Introduction}
Large language model (LLM)-based agentic systems have emerged as a promising paradigm for solving complex tasks by orchestrating multiple agents through natural language communication~\cite{wang2025evoagentx,wang2024survey,zhang2024aflow,tran2025multi}. 
Despite its human readability, natural language imposes fundamental constraints on inter-agent communication. To communicate, an agent must compress its rich, high-dimensional internal states into a sequence of discrete tokens, typically exposing only a single linear message (\ie a chain of thought (CoT)~\cite{wei2022chain} plan). This downsampling not only discards alternative reasoning paths, but also incurs substantial redundancy, as much of the generated text serves linguistic coherence rather than task-relevant information~\cite{zhang2024cut}. As a result, language-based communication can be ambiguous and lossy, which has been identified as a major source of coordination failures in multi-agent systems~\cite{chen2025reasoning,cemri2025multi}.

To move beyond language space, we explore the direct transmission of internal representations for more precise and information-preserving communication. In multi-agent settings, we refer to this as \textbf{\textit{latent communication}}.
While direct sharing is challenging for humans, which is often depicted in fictions~\cite{liu2008darkforest}, \ie \textit{telepathy}, LLM-based agents naturally perform most of their computation in latent space and produce rich hidden states throughout generation, which can be extracted to support direct, expressive communication. 
Previous hidden-state-based communication methods either rely on one-shot activation grafting~\cite{ramesh2025communicating} or remain coupled to language trajectories~\cite{tang2025augmenting}, and typically require ad-hoc layer selection, adding tuning overhead.

In this work, we propose \modelname, a novel framework that realizes this vision by enabling inter-agent communication entirely in latent space.
Rather than transmitting discrete tokens decoded by the language head, \modelname directly transmits the temporally aligned last-layer hidden states corresponding to an agent’s generated message, treating them as a representation of its thoughts. Under this formulation, we design a supervised objective that explicitly encourages the interpretation and utilization of task-relevant latent information, with a simple but effective stochastic token–latent mixing curriculum to stabilize training. To overcome the rigidity of full-trajectory message communication while preserving information integrity, we further train a separate reasoning process that autoregressively generates compact latent messages with a controllable number of generation steps directly in latent space, without decoding to language space tokens.
This allows \modelname to compress long reasoning trajectories into concise latent prefixes, substantially improving efficiency while retaining task-critical information for downstream execution.

Experimentally, we focus on a two-agent sender-receiver scenario, which is the building block of various multi-agent systems. To reduce confounding factors, we intentionally exclude components such as tool use, retrieval, or multi-round debate. 
Analysis reveals that agents utilizing latent communication exhibit more exploratory behavior patterns that lead to higher success rates by leveraging task-relevant latent information rather than superficial pattern matching. Moreover, we demonstrate that latent messages can be compressed to as few as \textbf{8 tokens} while maintaining competitive performance, achieving up to a $24\times$ reduction in communication latency. Further analysis of the output probability distribution after compression reveals how task-critical information is effectively preserved. 

\section{Related Work}\label{sec:related_work}
\paragraph{Latent Reasoning in LLMs.}
Recent research has begun shifting reasoning processes from the discrete language space to continuous latent representations, bypassing the bandwidth and efficiency limits of text ($\approx15$ bits/token vs. $\approx40$k bits/hidden-state)~\cite{zhu2025survey}. 
To expand computation during inference, \cite{goyal2023think} introduces pause tokens, while \cite{pfau2024let} employs filler tokens to scaffold intermediate reasoning. Beyond token scheduling, \cite{liu2024deliberation} proposes a latent coprocessor that modifies the transformer KV cache. Other work~\cite{hao2024training,shen2025codi,cheng2024compressed} enables multi-path parallel reasoning by feeding the last hidden state back as the next input embedding. Modular frameworks~\cite{bae2024relaxed,gao2024algoformer,geiping2025scaling} decouple encoding, latent reasoning, and decoding. 
Building upon these insights, we shift focus from \textit{single-model} latent reasoning to \textit{inter-agent} communication entirely in latent space.

\paragraph{Multi-agent Communication.}
LLM-based agent systems typically orchestrate in natural language~\cite{zhu2025multiagentbench,wang2024survey}, which can introduce ambiguity and computational overhead~\cite{zhang2024cut,yu2024natural,cemri2025multi,chen2025reasoning}.
Emergent communication studies~\cite{lazaridou2016multi,lazaridou2018emergence,tucker2021emergent,tucker2022trading} show that non-linguistic protocols can emerge, yet depend on channels learned from scratch and disconnected from internal reasoning.
Recent work explores richer forms:~\cite {pham2023let} transmits probability-weighted tokenizer embeddings;~\cite{ramesh2025communicating} blends hidden activations between agents; and~\cite{tang2025augmenting} communicates per-token latent deltas tied to language trajectories.\cite{zheng2025thought} infers latent “thoughts” from hidden states via an autoencoder and injects them as prefixes using recovered dependencies.
Unlike these works, our method directly transmits temporally aligned sequences of last hidden states and further compresses them to enable efficient, language-free communication.
\modelname preserves agent autonomy by requiring no parameter sharing~\cite{christianos2021scaling}, memory coupling~\cite{salemi2025llm}, or cache synchronization~\cite{fu2025cache,zou2025latent}, while expanding the effective communication bandwidth.

\section{\modelname}
In this section, we formalize how to extract an agent’s states as the representation of its \enquote{thought} for inter-agent latent space communication.
Let $x=(x_1,\dots,x_T)$ denote a sequence consisting of a prompt $x_{1:m}$ and a completion sequence $y=(y_1,\dots,y_L)$ such that $y_\ell = x_{m+\ell}$ and $L=T-m$.
For each decoding step $\ell=1,\dots,L$, define:
\begin{align*}
h_{\ell} &= \mathrm{Transformer}\!\left(x_{\le m+\ell-1}\right)_{m+\ell-1},\\
H &= [\,h_{1},\,h_{2},\,\dots,\,h_{L}\,],
\end{align*}
where $h_{\ell} \in \mathbb{R}^d$ is the last-layer hidden state immediately before predicting $y_\ell$ (\ie at position $m+\ell-1$ in the full sequence).
$H \in \mathbb{R}^{L\times d}$ collects these last hidden states for the completion region.

\begin{figure*}
    \centering
    \includegraphics[width=1\linewidth]{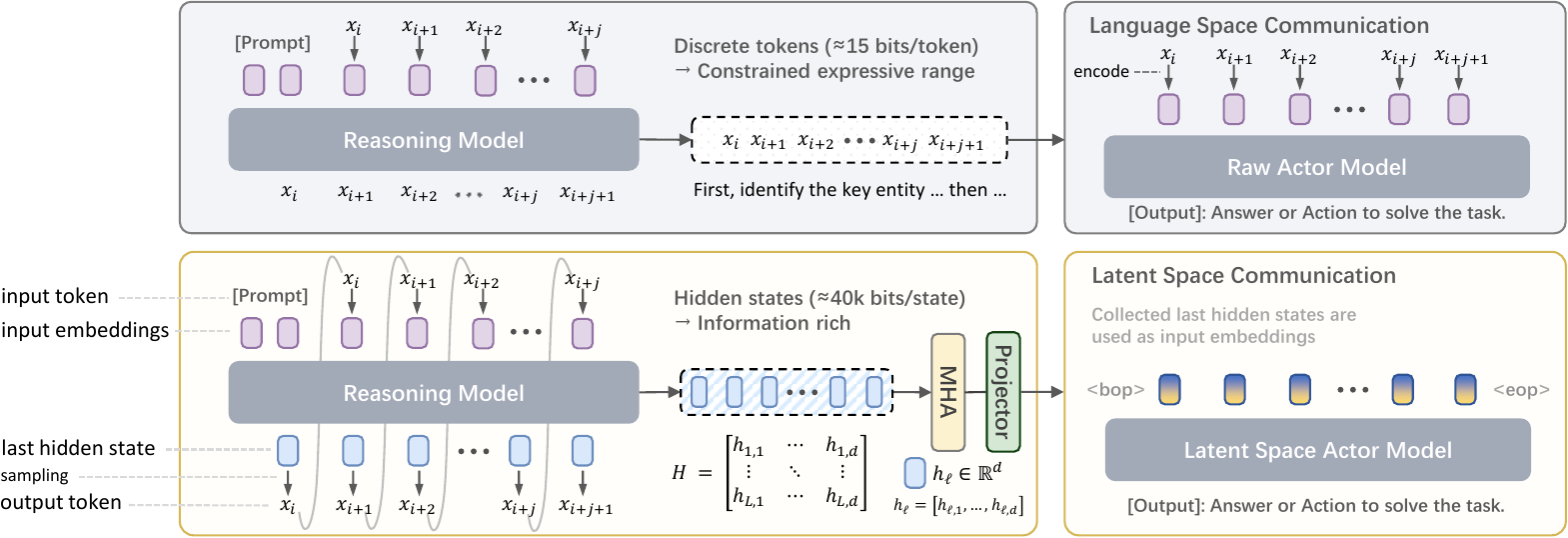}
    \caption{A comparison of \modelname with language-space communication. In language space, an agent transmits a discrete token sequence $[x_i, x_{i+1}, \dots, x_{i+j+1}]$ (\eg a CoT plan) to another. In \modelname, the model leverages its last hidden states as a representation of its internal \enquote{thought}, processed by a communication adapter, and then transmits them directly to the other agent, enabling communication entirely in latent space with higher expressive capacity.}
    \label{fig:architecture}
\end{figure*}

\subsection{Latent Communication}
\modelname removes natural language constraints by letting agents transmit their thoughts by directly passing the collected last hidden states, which we termed latent communication. As shown in Figure~\ref{fig:architecture}, this transmission occurs at the end of an agent's message generation process. Special tokens, $x_i=\texttt{<bop>}$ and $x_j=\texttt{<eop>}$, are added to mark the beginning and the end of the latent communications.
Consider an agent $\mathcal{M}_i$ solving a task $\mathcal{T}=\{x_1,\dots,x_m\}$.  
Upon receiving a latent communication $H = \{h_1, h_2, \dots, h_L\}$ from another agent, it forms its input embedding as:
\[
E \;=\; [\,e(x_1),\dots, e(x_i), h_1,h_2,\dots,h_L, e(x_j)\,],
\]
where $e(\cdot)$ is the token embedding. This inference process is analogous to language space multi-agent systems, except that it directly feeds hidden states between agents. The latent communications are processed by a trainable light-weight self-attention and a projection layer as a \textit{communication adapter} for rescaling and interpretation. 
\uline{For brevity, we may refer to \textit{latent communication} as \textit{latents} where unambiguous.}

\subsection{Training Procedure}
In this work, we consider two agents: a \textbf{\textit{reasoning}} agent as a sender that produces a task-specific plan together with its last-layer hidden states, and an \textbf{\textit{actor}} agent as a receiver that consumes this communication to generate actions to solve tasks. This two-agent setting can serve as a fundamental building block for more complex multi-agent systems.

Let $Y_t$ denote the next token at supervised position $t\in S$, where $S$ indexes the actor’s output tokens corresponding to the ground-truth response, and $C_t$ the decoder prefix up to position $t$.
We encourage the actor to utilize $H$ by maximizing a supervised fine-tuning objective regularized by conditional distributional separation:
\[
\mathcal{L}_{\mathrm{total}}
\;=\; \mathcal{L}_{\mathrm{task}}
\;+\; \lambda_{\mathrm{S}}\,\mathcal{L}_{\mathrm{sep}}
\;+\; \lambda_{\mathrm{A}}\,\mathcal{L}_{\mathrm{align}},
\]
where $\lambda_{\mathrm{S}}, \lambda_{\mathrm{A}} > 0$, and $\mathcal{L}_{\mathrm{task}}$ is the standard cross-entropy loss that ensures the model produces accurate and coherent responses for the given task.

\paragraph{Conditional thought separation.}
We compare the conditional output distributions $p_\theta$ induced by matched latent communications $H$ and mismatched latents $\tilde H$ (\ie latent communications sampled from a different task). Specifically, we minimize a weighted Jensen--Shannon divergence~\cite{lin2002divergence}:
\[
\mathcal{L}_{\mathrm{sep}}
= -\,\frac{1}{|S|}\sum_{t\in S}
\mathrm{JS}\!\big(p_\theta(\cdot \!\mid\! C_t,H),\, p_\theta(\cdot \!\mid\! C_t,\tilde H)\big).
\]
This objective separates matched from mismatched conditional distributions, providing a robust training signal that encourages the actor to attend to and leverage task-relevant latent information.

\paragraph{Plan-aligned regulation.}
While maximizing separation encourages sensitivity to $H$, it also introduces a failure mode where the model may exploit the objective by shifting probability mass toward idiosyncratic tokens that increase divergence while harming task utility.
To mitigate this, we regularize predictions conditioned on $H$ using those conditioned on the corresponding language-space plan $P$, generated by the same instruction-tuned model during the autoregressive generation of $H$.
Let $p_{\mathrm{plan}}(\cdot \mid C_t, P)$ denote the distribution when only the plan is provided. For brevity, we omit explicit conditioning on $C_t$ in the following.
\begin{align*}
&\mathcal{L}_{\mathrm{align}}
= \frac{\beta}{|S|}\sum_{t\in S}
    \mathrm{KL}\!\big(p_\theta(\cdot \mid H)\,\Vert\,
    p_{\mathrm{plan}}(\cdot \mid P)\big)
\notag\\
&+ \frac{\alpha}{|S|}\sum_{t\in S}
    \big(1-\cos\!\big(\mathbf{z}_\theta(H),\,
    \mathbf{z}_{\mathrm{plan}}(P)\big)\big),
\end{align*}
where $\mathbf{z}_\theta$ and $\mathbf{z}_{\mathrm{plan}}$ denote the corresponding normalized logit.
All divergences and cosine similarities are computed at supervised positions, with probabilities obtained from the softmax of logits.

\paragraph{Curriculum Learning.}
Learning to interpret latents from scratch is unstable. We thus adopt a token-to-latent curriculum that stochastically replaces early communication positions with their corresponding plan token embeddings. Concretely, we sample a replacement rate $r\in\{0,0.1,\dots,1.0\}$ and form a mixed communication
\[
H^{(r)} = \underbrace{e_1, \dots, e_{\lfloor r \cdot L \rfloor}}_{\text{token embeddings}} \oplus \underbrace{h_{\lfloor r \cdot L \rfloor + 1}, \dots, h_L}_{\text{latent states}}.
\]
This method enhances training efficiency while achieving strong model performance. 

\begin{figure}[t]
    \centering
    \includegraphics[width=1\linewidth]{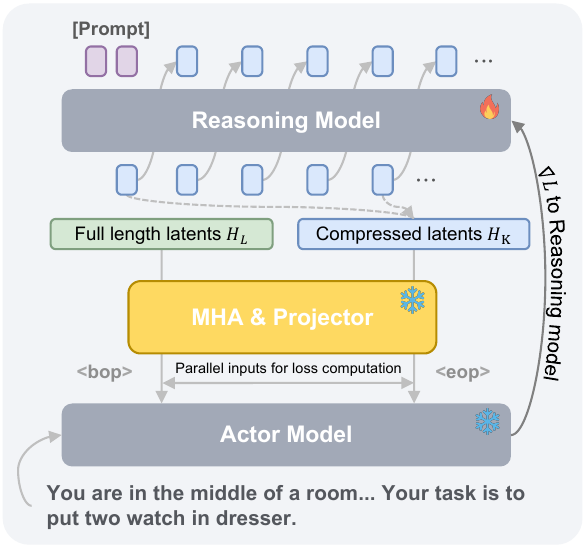}
    \caption{Training the reasoning model with frozen-actor supervision.}
    \label{fig:compress}
\end{figure}

\section{Information Compression}
While full-length latents $H_L \in \mathbb{R}^{L \times d}$ are highly expressive, their temporal length (often dozens to hundreds of steps) introduces substantial communication latency. Unlike natural-language tokens, whose semantics are discrete and inherently sequential, latent states are continuous and over-parameterized relative to task requirements, suggesting that much of the temporal structure is redundant. Our goal is therefore to learn an \textit{information-preserving bottleneck} that compresses latents while retaining their utility for downstream agents.

\paragraph{Compression via Latent-Space Reasoning.}
To this end, we train a separate reasoning model $M_\phi$ to generate compact latent messages $H_K \in \mathbb{R}^{K \times d}$ with $K \ll L$, while keeping the actor model and its communication adapter frozen. Rather than truncating or subsampling $H_L$, $M_\phi$ performs autoregressive reasoning entirely in latent space by feeding its last hidden state back as the next input embedding through a lightweight projection:
\[
\langle M_\phi(E_i) \rightarrow h_i,\; E_{i+1} = E_i \oplus \mathrm{Proj}(h_i) \rangle .
\]
This design enables an end-to-end differentiable latent generation loop without decoding to tokens, and isolates compression from changes in the actor’s behavior. During training, only the parameters of $M_\phi$ are updated, ensuring that compression is learned purely by adapting the latent message itself.

\paragraph{Training Objective.}
We train the compression model using a composite objective:
\[
\mathcal{L}_{\text{compress}} =
\lambda_{\text{task}} \mathcal{L}_{\text{task}} +
\lambda_{\text{pref}} \mathcal{L}_{\text{pref}} +
\lambda_{\text{geom}} \mathcal{L}_{\text{geom}},
\]
which jointly addresses the main failure modes of aggressive compression. The task loss $\mathcal{L}_{\text{task}}$ is a cross-entropy on the frozen actor’s predictions conditioned on $H_K$, ensuring downstream task utility.

Motivated by the non-uniform information density of token-level language communication~\cite{shannon1951prediction,zhang2024cut}, we introduce an uncertainty-weighted agreement loss that selectively aligns the actor’s behavior under compressed and full-length communications. Let $p_t^{(A)}$, $p_t^{(D)}$, and $p_t^{(B)}$ denote the actor’s output distributions at position $t$ when conditioned on $H_K$, $H_L$, and no latents, respectively. We define per-token weights
$
w_t \propto
\max\!\big(
\mathcal{H}(p_t^{(B)}) - \mathcal{H}(p_t^{(D)}),\, 0
\big)$.
where $\mathcal{H}(\cdot)$ denotes entropy, and compute
$\mathcal{L}_{\text{pref}}
= \sum_{t\in S} w_t \,
\mathrm{KL}\!\left(p_t^{(D)} \,\|\, p_t^{(A)}\right)$.
This objective emphasizes positions where latents meaningfully reduces predictive uncertainty, while avoiding over-regularization where latents are uninformative.

Finally, to prevent representational drift under strong compression, we apply a latent geometry alignment loss. Let $Z_k^{(A)}$ and $Z_k^{(D)}$ denote the actor-side latent features induced by $H_K$ and $H_L$ after adapter processing and length alignment, and define their step-averaged directions
$\bar{z}^{(A)}$ and $\bar{z}^{(D)}$. We enforce $\mathcal{L}_{\text{geom}}
= 1 - \cos\!\left(
\bar{z}^{(A)},\, \bar{z}^{(D)}
\right)$,
which preserves the global semantic orientation of the original communication in the compressed latent space. Together, these objectives encourage $M_\phi$ to learn an information-preserving bottleneck that discards redundant temporal structure while retaining task-critical functional and geometric properties.
Full derivations are provided in Appendix~\ref{sec:compression}.

\begin{table*}[t]
\centering
\renewcommand{\arraystretch}{0.8}
\resizebox{\textwidth}{!}{
\begin{tabular}{lrrrrrrrrrrrr}
\toprule[1.5pt]
\multicolumn{1}{c}{\multirow{2}{*}{Method}} 
& \multicolumn{4}{c}{\rule{0pt}{2.4ex}Qwen2.5-7B-Base} 
& \multicolumn{4}{c}{Qwen2.5-0.5B-Base}
& \multicolumn{4}{c}{LLaMA3.1-8B} \\
\cmidrule(lr){2-5}\cmidrule(lr){6-9}\cmidrule(lr){10-13}
\multicolumn{1}{c}{} 
& Seen & Steps & Unseen & Steps 
& Seen & Steps & Unseen & Steps
& Seen & Steps & Unseen & Steps \\
\toprule[1pt]
\textbf{\modelname} \\
\toprule[0.25pt]
\rowcolor{lightpurple}Ours 
& \textbf{70.48} & 9.41/12.54 & \textbf{65.42}  & 9.86/13.37
& \textbf{61.19} & 10.55/14.22 & \textbf{57.46} & 9.38/13.90
& \textbf{70.71} & 8.02/12.58 & \textbf{70.90} & 8.21/12.96 \\
Text 
& 64.29 & 8.76/12.77 & 62.44 & 9.79/13.63
& 54.52 & 9.50/14.28 & 47.26 & 9.70/15.13
& 62.86 & 7.91/12.94 & 60.82 & 8.14/13.21 \\
No-Comm 
& 62.14 & 10.19/13.90 & 62.19 & 10.23/13.92
& 50.48 & 8.23/14.06 & 44.03 & 9.10/15.20
& 63.57 & 8.35/12.59 & 58.40 & 9.47/13.85 \\
\toprule[1pt]
\textbf{Baselines} \\
\toprule[0.25pt]
CoT (full) 
& \underline{67.14} & 8.15/12.04 & \underline{64.93} & 9.02/12.87
& \underline{57.86} & 8.30/13.23 & 50.75 & 8.94/14.39
& \uline{69.35} & 7.62/12.32 & \uline{70.82} & 7.88/12.47 \\
No-CoT 
& 65.71 & 8.23/12.27 & 62.69 & 9.15/13.20
& 57.14 & 8.96/13.69 & 50.25 & 9.80/14.87
& 67.18 & 7.85/12.61 & 70.34 & 8.02/12.88 \\
\toprule[1pt]
\textbf{Variants} \\
\midrule[0.25pt]
CrossTask 
& 61.43 & 8.42/12.89 & 61.94 & 9.51/13.50
& 53.57 & 9.40/14.32 & 47.01 & 10.06/15.33
& 65.00 & 8.05/12.24 & 63.43 & 9.86/13.57 \\
Noised & & & & & & & & & & & & \\
\quad CovNoise-0.5$\times$ 
& 64.29 & 8.54/12.63 & 60.95 & 8.71/13.12
& 53.33 & 8.80/14.03 & 46.77 & 9.64/15.16
& 64.29 & 8.10/12.50 & 65.68 & 9.47/12.77 \\
\quad CovNoise-1.0$\times$ 
& 63.81 & 8.66/12.76 & 63.68 & 8.72/12.82
& 53.10 & 8.96/14.14 & 44.53 & 9.68/15.40
& 58.57 & 7.80/12.80 & 64.93 & 9.66/13.28 \\
\quad WhiteNoise 
& 61.90 & 8.65/12.97 & 61.19 & 9.32/13.46
& 57.38 & 8.00/13.11 & \underline{57.21} & 9.18/13.81
& 61.43 & 8.01/12.64 & 64.93 & 9.52/13.19 \\
CovGauss-0$\mu$ 
& 60.00 & 8.79/13.27 & 61.94 & 9.59/13.55
& 13.81 & 11.25/18.79 & 13.18 & 12.93/19.07
& 57.86 & 8.04/13.08 & 66.42 & 9.51/13.03 \\
CovGauss-$\mu$ 
& \underline{65.71} & 8.58/12.50 & \underline{64.93} & 8.63/12.62
& 44.52 & 9.21/15.20 & 34.33 & 10.19/16.63
& 60.71 & 7.69/12.53 & 64.93 & 8.85/12.76 \\
RandomRot 
& 57.86 & 8.43/13.31 & 63.68 & 9.37/13.23
& 59.05 & 8.24/13.06 & 51.99 & 9.12/14.34
& 57.86 & 7.67/12.86 & 63.44 & 9.04/13.25 \\
\toprule[1pt]
\textbf{Cross family} \\
\toprule[0.25pt]
Qwen2LLaMA & 70.95 & 8.47/12.01 & 71.39 & 9.21/13.05 & – & – & – & – & – & – & – & –\\
\toprule[1.5pt]
\end{tabular}
}
\caption{Performance of different methods and variants on seen and unseen tasks of Alfworld under three model backbones. Higher success rates indicate stronger inter-agent collaboration and task-solving ability. “Steps” reports average steps on successful tasks and average steps over all tasks. Best in \textbf{bold}, second-best \underline{underlined}.}
\label{tab:overall}
\end{table*}

\begin{table}[t]
    \centering
    \renewcommand{\arraystretch}{0.85}
    \setlength{\tabcolsep}{6pt}
    \footnotesize
    \begin{tabular}{lcccc}
    \toprule[1pt]
    Method & Overall & Level-3 & Level-4 & Level-5 \\
    \midrule[0.5pt]
    \rowcolor{lightpurple}Ours        & \underline{36.88} & 40.08 & \underline{27.45} & \textbf{15.80} \\
    Text        & 34.35 & 37.60 & 26.30 & 14.20 \\
    No-Comm     & 33.27 & 36.40 & 26.20 & 13.10 \\
    \midrule[0.35pt]
    CoT (full)  & \textbf{38.35} & \textbf{45.65} & \textbf{31.19} & 15.05 \\
    No-CoT      & 36.25 & \underline{40.10} & 26.80 & 14.80 \\
    \bottomrule[1pt]
    \end{tabular}
    \caption{Accuracy (\%) on the MATH benchmark. While CoT benefits from linguistic constraints on simpler tasks, \modelname outperforms the strong CoT baseline on the most challenging Level-5 tasks.}
    \label{tab:math}
\end{table}

\section{Experiments}
\paragraph{Implementation Details.}
We evaluate our approach on Alfworld~\citep{shridhar2020alfworld}, and MATH~\cite{hendrycks2021measuring}. For Alfworld, training is conducted using data from~\cite{song2024trial}.
Qwen2.5-7B/0.5B-Base~\cite{qwen2.5} and LLaMA3.1-8B-Base~\cite{dubey2024llama} are employed as actor agents to isolate benefits from instruction-tuning priors. 
CoT plans and compression-free latents are generated by their instruction-tuned counterparts. 
We use base models as reasoning models in the information compression experiments. 
Alfworld episodes are capped at 20 steps; unfinished episodes are failures.
All models are trained using bfloat16, FlashAttention-2~\cite{dao2023flashattention}, and DeepSpeed~\cite{rajbhandari2020zero}. For the actor model, we optimize using AdamW~\cite{loshchilov2017decoupled} with a learning rate of $1\times 10^{-5}$, a global batch size of 16, and a 3\% linear warmup. We fix $\lambda_{\text{task}} = 1$ and dynamically anneal the regularization coefficients during training ($\lambda_{\text{sep}} \in [0.1, 1.0]$, $\lambda_{\text{align}} \in [0.1, 0.2]$). Negative samples for $\mathcal{L}_{\text{sep}}$ are constructed using latents from different tasks within the same batch. For information compression, we train the reasoning agent with a frozen actor, utilizing a learning rate of $5\times 10^{-5}$ and unit weights for all three compression objectives. We select the best models based on a 5\% validation split. All reported results are averaged over three independent runs. 

\paragraph{Baselines and variants in \modelname.}
We study the feasibility of \modelname and compare against two baselines:
\textbf{CoT (full)} uses complete CoT plans from instruction-tuned models for full-parameter supervised fine-tuning; 
\textbf{No-CoT} directly predicts final answers without any plan.

We further evaluate \textit{variants} of our method:
\textbf{Text} replaces latent messages with the corresponding CoT plan; 
\textbf{No-Comm} removes communication entirely; 
\textbf{CrossTask} replaces the current task’s latents with one sampled from a different task.
\textbf{Noised} adds structured or unstructured perturbations to $H$; 
\textbf{CovGauss} and \textbf{RandomRot} preserve mean or covariance statistics while destroying higher-order structure. \textbf{Qwen2LLaMA} uses latents from Qwen2.5-7B to train LLaMA3.1-8B model.
See Appendix~\ref{app:baseline} for detailed implementation setups.

\subsection{Main Results}
Table~\ref{tab:overall} presents a comprehensive comparison of the \modelname framework against baselines. Latent communication improves agents’ task-solving performance, as evidenced by gains over both fine-tuned single-agent baselines and agents trained to communicate in natural language. We highlight several key observations below.

\paragraph{Latent Communication Prompts Exploration.}
Beyond improvements in success rates, latent communication enables agents to execute longer yet more successful trajectories. By leveraging multiple plausible reasoning paths encoded in latents from other agents, the actor naturally exhibits more thorough exploratory behavior, even without explicit exploration training. Importantly, this increased trajectory length correlates with higher success rates rather than degraded efficiency, indicating informed exploration instead of random wandering. This behavior suggests a stronger environmental understanding enabled by latent communication, where parallel hypotheses are preserved and gradually resolved during action execution. This pattern is analyzed in detail in Appendix~\ref{app:ablation}.

\begin{figure}[t]
    \centering
    \includegraphics[width=0.98\linewidth]{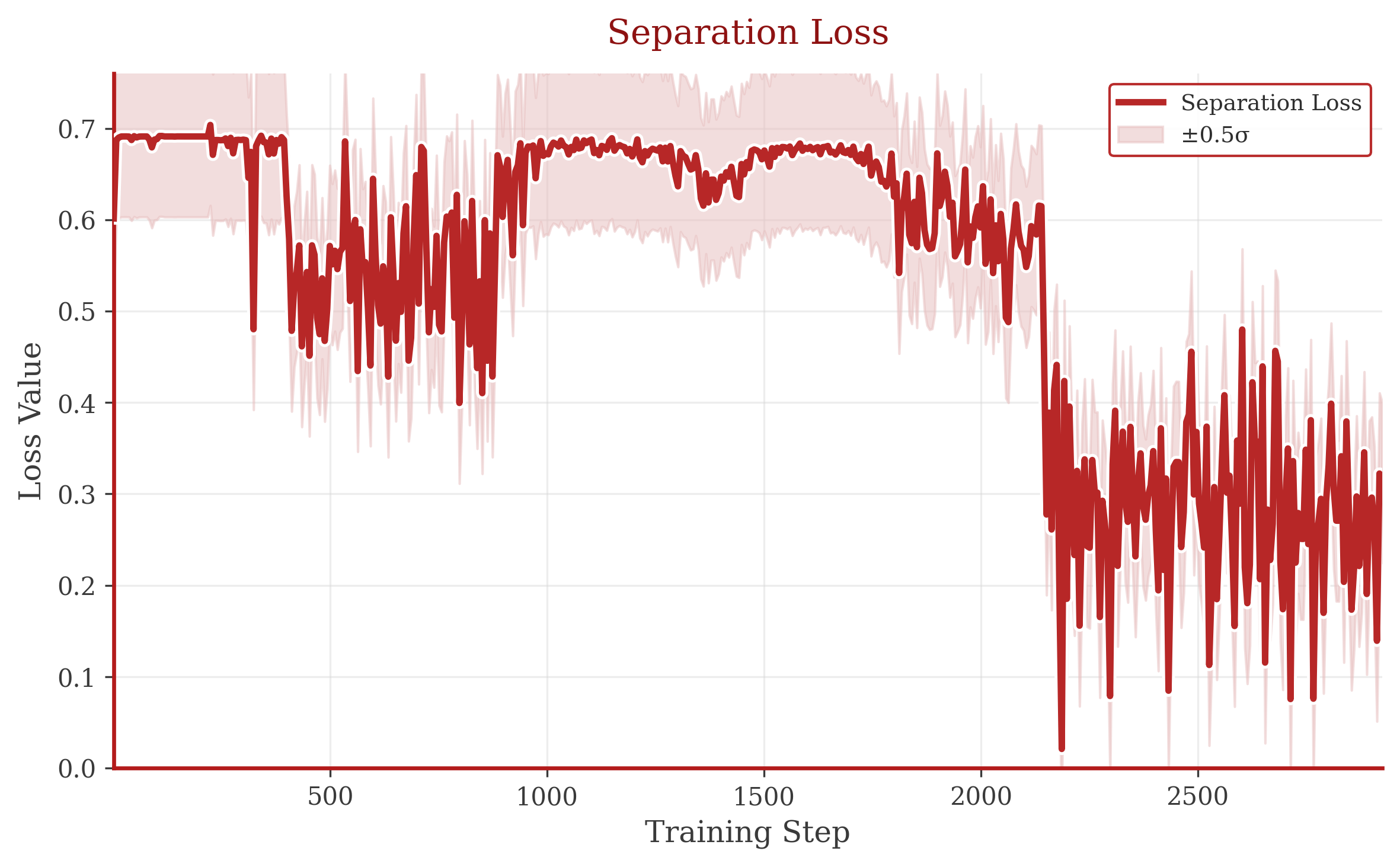}
    \caption{Training dynamics of the separation loss: an initial plateau near 0.69 indicates no separation between matched/mismatched latents, followed by a sharp drop after $\sim2.2$k steps, marking the model’s “aha” moment in exploiting task-relevant latent information.}
    \label{fig:dual-loss}
\end{figure}

\begin{table*}[htbp]
\centering
\small
\begin{minipage}{.48\linewidth}
\centering
\begin{tabular}{llll}
\toprule
Ratio & Seen & Unseen & Time \\
\midrule
\multicolumn{4}{l}{\textbf{Untrained (An instruction-tuned Model)}} \\
\midrule
Full  & $\underline{70.48}_{\pm 1.01}$ & $\underline{65.42}_{\pm 0.87}$ & 9.19s \\
90\%  & $68.57_{\pm 1.63}$             & $\textbf{67.16}_{\pm 1.97}$     & - \\
80\%  & $68.10_{\pm 1.83}$             & $61.69_{\pm 1.43}$              & - \\
70\%  & $67.14_{\pm 1.82}$             & $63.43_{\pm 2.24}$              & - \\
60\%  & $66.43_{\pm 1.63}$             & $59.20_{\pm 3.69}$              & - \\
50\%  & $\textbf{72.14}_{\pm 1.48}$    & $61.19_{\pm 2.84}$              & - \\
40\%  & $66.90_{\pm 2.31}$             & $59.95_{\pm 2.64}$              & - \\
30\%  & $65.95_{\pm 2.12}$             & $62.19_{\pm 1.58}$              & - \\
20\%  & $67.86_{\pm 3.23}$             & $61.44_{\pm 1.58}$              & - \\
10\%  & $67.86_{\pm 2.12}$             & $62.44_{\pm 2.64}$              & - \\
5\%   & $64.29_{\pm 1.12}$             & $60.95_{\pm 1.35}$              & - \\
0\%   & $62.14_{\pm 2.01}$                        & $62.14_{\pm 2.32}$                         & - \\
\bottomrule
\end{tabular}
\end{minipage}\hfill
\begin{minipage}{.48\linewidth}
\centering
\begin{tabular}{llll}
\toprule
Ratio & Seen & Unseen & Time \\
\midrule
\multicolumn{4}{l}{\textbf{Untrained (An instruction-tuned Model)}} \\
\midrule
128L & $64.55_{\pm 2.26}$ & $60.25_{\pm 2.06}$ & 3.55s \\
64L  & $66.23_{\pm 1.95}$ & $61.53_{\pm 4.32}$ & 1.83s \\
32L  & $63.57_{\pm 2.01}$ & $60.18_{\pm 3.58}$ & 1.03s \\
16L  & $64.29_{\pm 1.34}$ & $60.00_{\pm 3.01}$ & 0.62s \\
8L   & $64.00_{\pm 2.18}$ & $57.46_{\pm 2.69}$ & 0.39s \\
\midrule
\multicolumn{4}{l}{\textbf{Trained}} \\
\midrule
128L & $\textbf{68.10}_{\pm 1.93}$ & $\textbf{62.94}_{\pm 2.03}$ & 2.25s \\
64L  & $\underline{67.14}_{\pm 1.56}$ & $\underline{61.94}_{\pm 2.13}$ & 1.16s \\
32L  & $66.90_{\pm 1.46}$ & $61.94_{\pm 2.56}$ & 0.60s \\
16L  & $66.43_{\pm 2.05}$ & $61.69_{\pm 2.56}$ & 0.33s \\
8L   & $66.43_{\pm 1.22}$ & $60.45_{\pm 2.23}$ & 0.20s \\
\bottomrule
\end{tabular}
\end{minipage}
\caption{Compression results on Alfworld.
\textbf{Left:} training-free sweep over retained ratio $R$.
\textbf{Right:} varying latent length with untrained and trained models.
Time denotes end-to-end latency (s) of the message generation process.
Best results are \textbf{bold}, second-best are \underline{underlined}. \textbf{See Appendix~\ref{app:compression} for results on LLaMA model.}}
\label{tab:compression}
\end{table*}

\begin{figure}[t]
    \centering
    \includegraphics[width=1\linewidth]{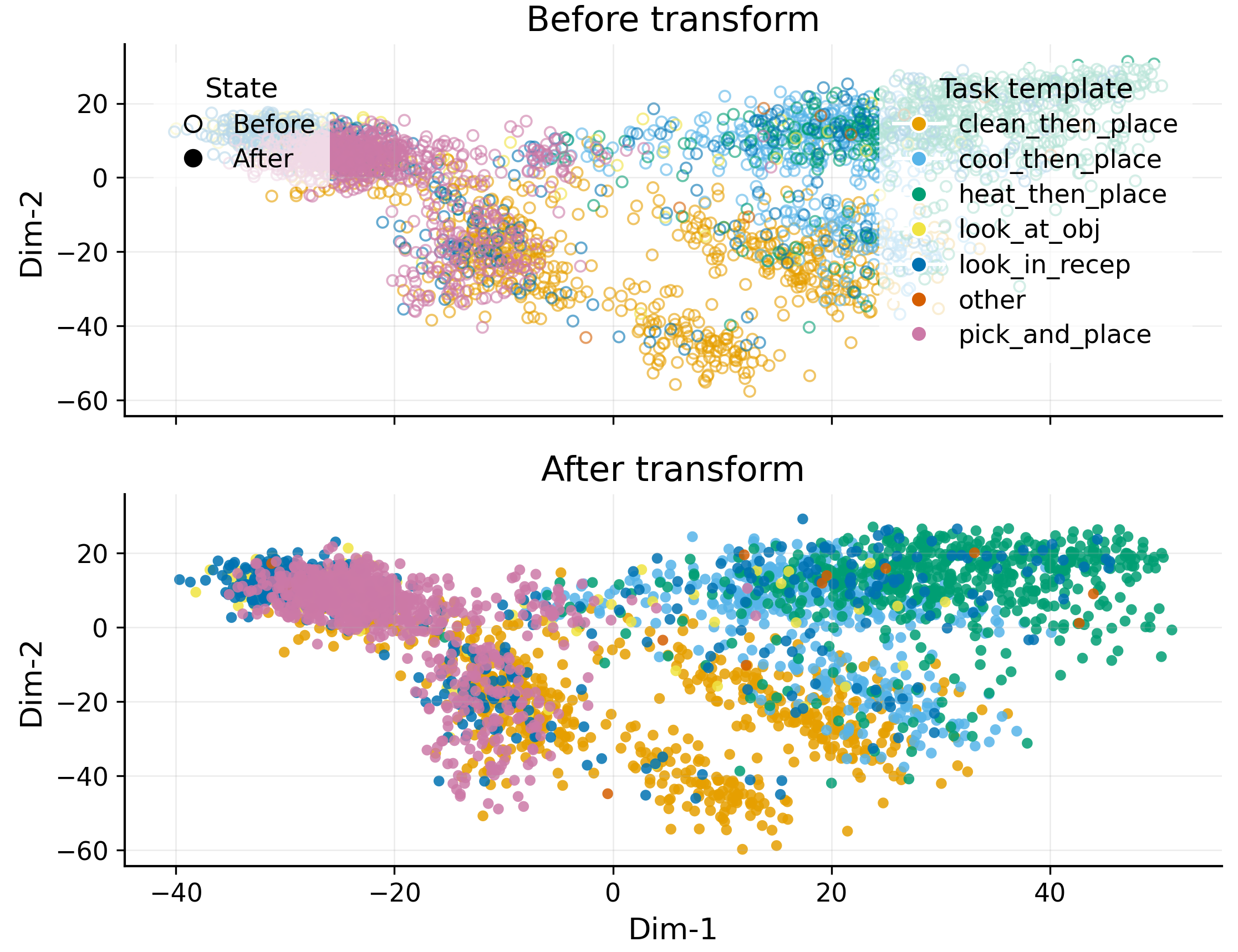}
    \caption{PCA visualization of latent communications, showing distinguishable task-specific structure in latent space both before and after the communication adapter.}
    \label{fig:pca}
\end{figure}

\paragraph{Semantics and Learning Dynamics.}To assess whether the actor genuinely exploits latent information rather than superficial patterns, we conduct structured perturbation experiments. Replacing task-matched latents with cross-task latents leads to a substantial performance drop, indicating that the actor relies on task-specific reasoning content encoded in the latents. Performance degrades further under covariance-matched Gaussian surrogates or random orthogonal rotations, which preserve first- and second-order statistics while destroying higher-order structure, supporting the interpretation that the actor is sensitive to meaningful latent geometry rather than low-order moments alone. Additive and white noise perturbations similarly impair performance, further indicating reliance on structured internal information instead of noise-robust heuristics.
Experiments involving \textit{cross-family} latent inputs: feeding Qwen-derived latents to train an LLaMA actor, yield even stronger performance gains. Since these model families exhibit distinct latent manifolds, the improvement cannot be attributed to superficial architectural compatibility. Instead, it suggests latent-level inter-agent understanding that transfers across heterogeneous representations. This observation aligns with findings in language-space agentic systems, where heterogeneous LLM agents often outperform homogeneous ensembles due to complementary inductive biases and reduced error correlations~\cite{shinn2023reflexion,wu2024autogen}.To corroborate these findings qualitatively, Figure~\ref{fig:pca} and Appendix~\ref{app:pca} visualize clear semantic clustering before and after processing by the communication adapter, confirming effective semantic alignment for downstream use.

Training dynamics further reveal how the actor learns to interpret latent communication. As shown in Figure~\ref{fig:dual-loss}, the separation loss remains near $\ln 2$ for approximately the first 2k steps, indicating no effective distinction between matched and mismatched messages. It then drops sharply, marking an ``aha'' moment in which the actor begins to exploit and leverage task-relevant latents, consistent with the intended effect of the separation objective.

\paragraph{Generalization to Symbolic Reasoning.}
To assess generalization beyond interactive settings, we evaluate \modelname on the MATH benchmark. While prior work on latent-space reasoning~\cite{hao2024training,shen2025codi,ramesh2025communicating} often reports degraded performance relative to CoT supervision, Table~\ref{tab:math} reveals an intriguing inversion: although \modelname slightly underperforms on simpler problems, it surpasses the CoT baseline on the most challenging Level~5 tasks. We attribute this to the duality of linguistic constraints. For lower-complexity problems, the strict linearization of natural language acts as a beneficial regularizer, efficiently pruning the search space. However, for high-complexity tasks, this forced discretization causes a ``premature collapse'' of the reasoning distribution. In contrast, \modelname maintains a superposition of parallel hypotheses in its continuous representations. This capability allows the model to effectively conduct a broader search in latent space that is inaccessible to linear text decoding.

\begin{figure*}[t]
    \centering
    \includegraphics[width=\linewidth]{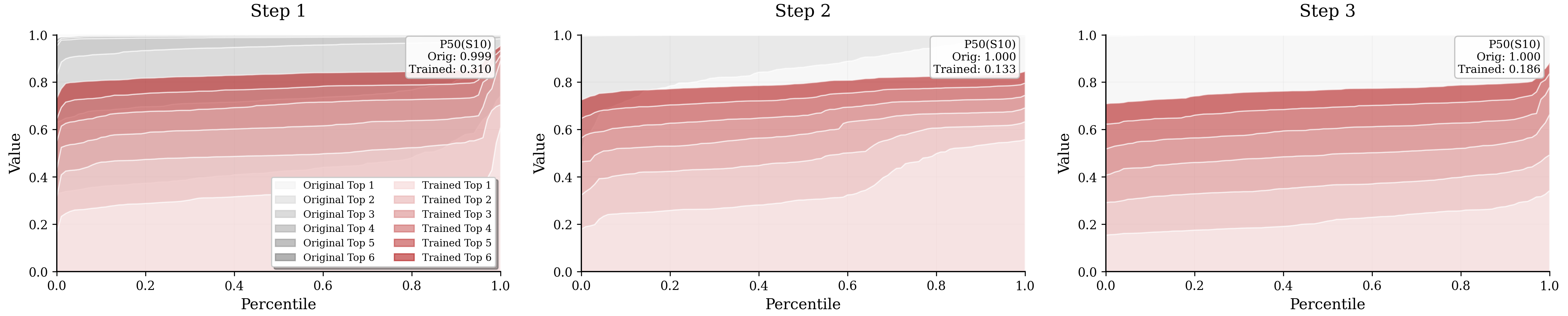}
    \caption{Parallelism analysis of latent communication over the first three steps.
    Trained latents (red) maintain stable Top-$k$ gaps and lower $P_{50}(S_{10})$, indicating stronger parallelism, while untrained latents (gray) collapse toward Top-1.
    See Appendix~\ref{sec:parallelism} for extended results.}
    \label{fig:topk-6}
\end{figure*}

\subsection{Compression Analysis}

\paragraph{Compression Performance.}
Theoretically, due to their substantially higher expressive capacity, latent communications can encode rich information in far fewer positions. 
To quantify this compression capacity, we consider two settings. \textbf{i) Untrained:} We directly use Qwen2.5-7B-Instruct to generate full-length latents for actor training, and then truncate them to shorter lengths. This setting evaluates the empirical compressibility of raw latents. \textbf{ii) Trained:} We use a compression-trained Qwen2.5-7B-Base reasoning model. Results on the LLaMA model are provided in Appendix~\ref{app:compression}.

As shown in Table~\ref{tab:compression}, naive truncation performs best at moderate compression (50\%) but degrades under more aggressive shortening, revealing the limits of untrained compression.
In contrast, compression training enables consistently higher and more stable success rates across compressed latent ranging from 8 to 128 steps (around 1.8\% to 28.8\% of the full sequence), indicating that the reasoning model learns an information-preserving pattern that discards temporal redundancy while retaining task-relevant semantics. Furthermore, compression substantially improves efficiency, reducing end-to-end latency from 9.19\,s to 0.39\,s with 8-step latents (nearly $24\times$ speed-up), and further to 0.20\,s with a lightweight bridge module by largely eliminating decode--re-encode overhead.

\begin{figure}[t]
    \centering
    \includegraphics[width=1\linewidth]{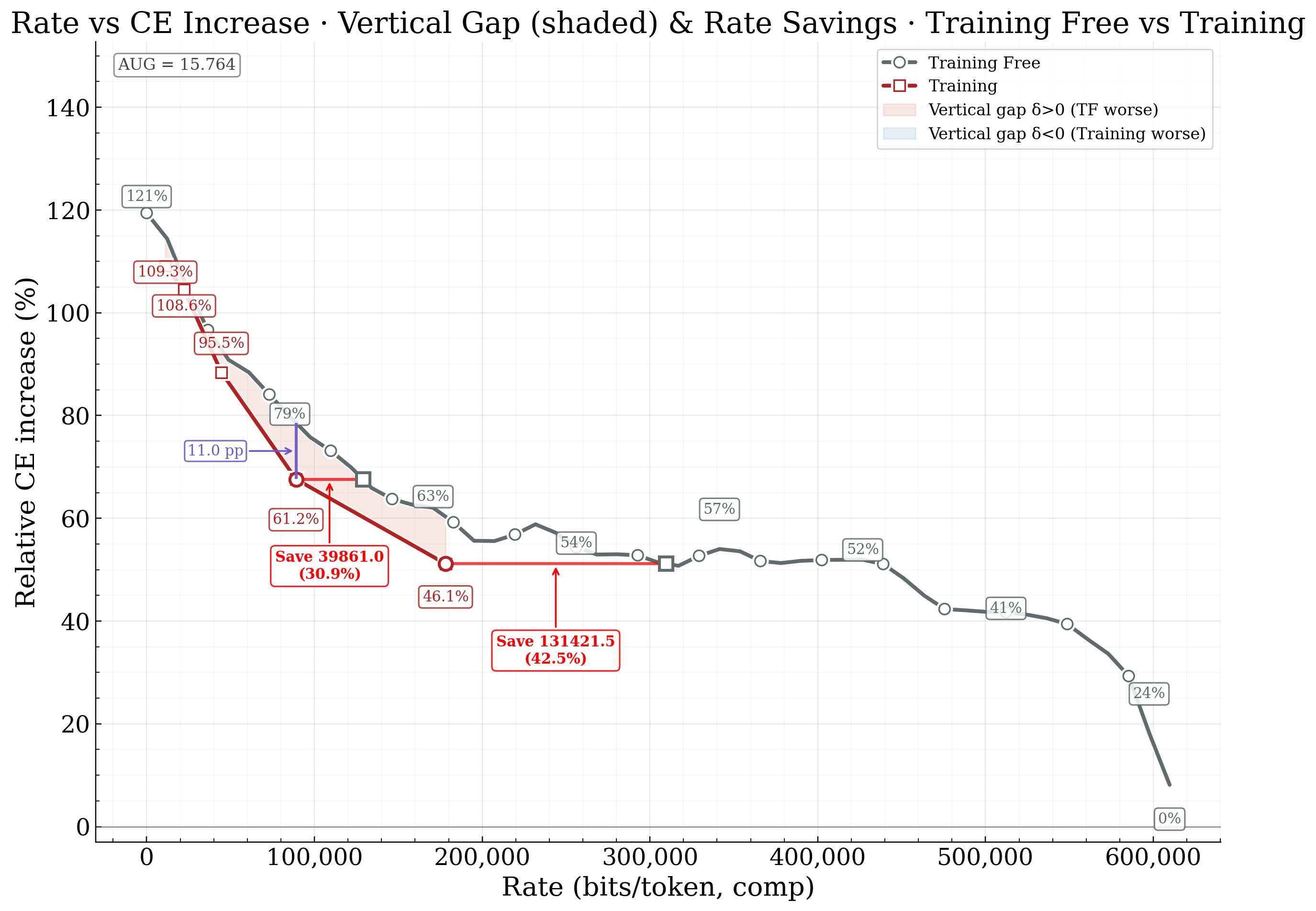}
    \caption{Task-averaged relative change $\Delta\mathrm{CE}$ and relative savings before and after compression training.}
    \label{fig:rate_ce}
\end{figure}

\paragraph{Why compression is effective.}\label{compress_effective}
To understand why compression preserves performance, we analyze its effect on the actor’s predictive uncertainty.
We sweep the communication rate $R \in [0,1]$ and measure the task-averaged relative change in cross-entropy (CE),
$\Delta\mathrm{CE}\%(R) = 100 \times \frac{\mathrm{CE}_{\mathrm{comp}}(R) - \mathrm{CE}_{\mathrm{full}}}{\mathrm{CE}_{\mathrm{full}}}$.
As shown in Figure~\ref{fig:rate_ce}, $\Delta\mathrm{CE}\%$ decreases monotonically with increasing $R$ and plateaus between roughly 30\% and 75\%, aligning with the range of strongest empirical performance.
Across all rates, learned compression consistently yields lower CE than training-free truncation, with a maximum gap of approximately 11 percentage points, indicating better preservation of predictive confidence under reduced communication.

We further examine how information is preserved under compression via the actor’s output distributions. Following~\cite{hao2024training}, broader probability mass is interpreted as indicating that the model maintains more plausible alternatives.
Figure~\ref{fig:topk-6} analyzes the latent steps of the reasoning agent by plotting the cumulative probability mass of the top-$6$ tokens across communication percentiles, with probabilities normalized over the top-10 tokens for comparability. Latents produced by the trained reasoning agent exhibit stable gaps between successive top-$6$ curves across steps, whereas untrained compression shows rapid concentration toward top-ranked tokens. We quantify this behavior using a head-coverage statistic, $\mathrm{P50}(S_{10})$, defined as the median cumulative probability mass of the top-10 tokens, which is consistently lower for the trained model and indicates broader support over plausible alternatives.

Together, these observations suggest that learned compression preserves diverse hypotheses across multiple reasoning steps, avoiding collapse into a single trajectory and thereby retaining richer information for downstream decision making.

\begin{table*}[htp]
\centering

\setlength{\tabcolsep}{8pt}
\renewcommand{\arraystretch}{1.05}

\begin{minipage}{0.49\linewidth}
\centering
\begin{tabular}{lcc}
\toprule
\textbf{Actor} & \textbf{Seen} & \textbf{Unseen} \\
\midrule[\cmidrulewidth] 
Ours Full & $\textbf{70.48}_{\pm 1.01}$ & $\textbf{65.42}_{\pm 0.87}$ \\
\cmidrule(lr){1-3}       
$\text{w/o}$ curri & $33.10_{\pm 2.97}$ & $20.65_{\pm 2.15}$ \\
$\text{w/o}$ $\mathcal{L}_\mathrm{sep}$ & $\underline{58.81}_{\pm 1.41}$ & $\underline{60.70}_{\pm 5.50}$ \\
$\text{w/o}$ $\mathcal{L}_\mathrm{align}$ & $56.90_{\pm 1.41}$ & $53.98_{\pm 3.35}$ \\
$\text{w/o}$ adapter & $4.05_{\pm 1.70}$ & $4.48_{\pm 1.31}$ \\
\bottomrule
\end{tabular}
\end{minipage}
\hfill
\begin{minipage}{0.49\linewidth}
\centering
\begin{tabular}{lcc}
\toprule
\textbf{Reasoning} & \textbf{Seen} & \textbf{Unseen} \\
\midrule[\cmidrulewidth]
Ours Full & $\textbf{68.10}_{\pm 1.93}$ & $\underline{62.94}_{\pm 2.03}$ \\
\cmidrule(lr){1-3}
$\text{w/o}$ $\mathcal{L}_{\mathrm{task}}$ & $65.71_{\pm 1.43}$ & $\textbf{63.18}_{\pm 3.47}$ \\
$\text{w/o}$ $\mathcal{L}_\mathrm{pref}$ & $64.76_{\pm 2.97}$ & $60.20_{\pm 3.13}$ \\
$\text{w/o}$ $\mathcal{L}_\mathrm{geom}$ & $64.05_{\pm 3.55}$ & $59.45_{\pm 3.01}$ \\
\bottomrule
\end{tabular}
\end{minipage}
\caption{Ablation of training components. \textbf{Left:} For the actor, the communication adapter and curriculum learning are foundational; removing the adapter leads to near-zero success rates. \textbf{Right:} For the reasoning model, the geometry alignment loss ($\mathcal{L}_{\mathrm{geom}}$) proves to be the most critical objective for maintaining performance under compression. \textbf{See Appendix~\ref{app:ablation} for full results and analysis.}}
\label{tab:ablation}
\end{table*}

\subsection{Ablation Studies}
Table~\ref{tab:ablation} presents a systematic ablation of the training components. For the actor model, removing curriculum learning forces the model to interpret latents from scratch, leading to extremely unstable training dynamics and severely degraded comprehension (Appendix, Figure~\ref{fig:ablation_curriculum}). Removing the separation loss induces shortcut behavior; the model learns to ignore the latent communication and rely only on the textual task prompt, causing performance to regress toward the single agent baseline. Removing the communication adapter causes the largest drop; despite generating fluent and coherent responses, the model fails to complete tasks, underscoring the adapter’s role in bridging the agents’ latent spaces and enabling interpretation of latents.

For the reasoning model, which is trained to generate compressed latents, we ablated its three core loss functions with compressed target length $K=128$. The most critical component is the direction alignment loss ($\mathcal{L}_\mathrm{geom}$). This highlights the importance of maintaining geometric consistency between the compressed latents and the uncompressed ones. The agreement loss ($\mathcal{L}_\mathrm{pref}$) is also vital, as removing it significantly impairs the model's ability to produce latents that elicit the correct behavior from the actor. Removing the cross-entropy loss ($\mathcal{L}_\mathrm{task}$) degrades performance on seen tasks but slightly improves unseen performance, suggesting a minor trade-off between in-distribution optimization and generalization. We leave a deeper investigation into this trade-off to future work. 

\begin{table*}[htbp]
    \centering
    \resizebox{0.95\textwidth}{!}{
    \begin{tabular}{@{}llccl@{}}
        \toprule
        \textbf{Communication} & \textbf{Topology ($N$)} & \textbf{Seen (\%)} & \textbf{Unseen (\%)} & \textbf{Structural Description} \\
        \midrule
        \multirow{3}{*}{\textit{Text (CoT)}} 
        & text-chain-2 & 57.86 & 50.75 & Sequential: 1 Sender $\rightarrow$ 1 Actor \\
        & text-chain-3 & 58.34 & 51.89 & Sequential: Sender $\rightarrow$ Sender $\rightarrow$ Actor \\
        & text-tree-3  & 61.19 & 56.75 & Parallel: 2 Senders $\rightarrow$ Actor (Merged by agent) \\
        \midrule
        \multirow{3}{*}{\textit{Latent (Ours)}} 
        & chain-2 (Base) & 61.19 & 57.46 & Sequential: 1 Sender $\rightarrow$ 1 Actor \\
        & chain-3 & 62.14 & 59.44 & Sequential: Sender $\rightarrow$ Sender $\rightarrow$ Actor \\
        & tree-3  & \textbf{62.85} & \textbf{60.77} & Parallel: 2 Senders $\rightarrow$ Actor (Adapter merged) \\
        \bottomrule
    \end{tabular}
    }
    \caption{Performance of complex agent topologies on ALFWorld using Qwen2.5-0.5B-Base. We compare natural language CoT communication (\enquote{text-}) with Interlat (\enquote{chain-} and \enquote{tree-}) across varying agent counts ($N$). Latent communication consistently yields higher success rates, with the parallel tree topology performing the best.}
    \label{tab:complex_topologies}
\end{table*}

\subsection{Generalization to Complex Topologies.} 
To directly verify that Interlat scales beyond simple pairwise communication, we evaluate it on 3-agent topologies using the Qwen2.5-0.5B-Base model on the ALFWorld benchmark. We ensure the entire multi-agent system can be jointly trained end-to-end by having all non-output sender agents generate fixed-length ($N=32$) messages entirely in latent space. Specifically, we investigate two distinct configurations:

\textbf{Sequential Chain Topology.} In this sequential structure (Agent 1 $\rightarrow$ Agent 2 $\rightarrow$ Actor), a \textit{Strategist} agent first generates a high-level latent plan capturing the task goals. A \textit{Compiler} agent then refines this received plan into a more actionable representation before passing it to the final Actor. 

\textbf{Parallel Tree Topology.} This parallel structure ([Agent 1 + Agent 2] $\rightarrow$ Actor) utilizes an \textit{Explorer} agent to generate diverse candidate reasoning paths and a \textit{Critic} agent to produce verification constraints simultaneously. To dynamically fuse these parallel messages, we extend the communication adapter with an additional learned projector that maps the aggregated latents from multiple senders into a unified representation for the Actor.

As shown in Table \ref{tab:complex_topologies}, both 3-agent latent configurations consistently outperform their natural language CoT counterparts with identical agent structures (e.g., 62.85\% vs. 61.19\% on seen tasks for the parallel tree). Moreover, both configurations yield solid improvements over the original 2-agent Interlat baseline (61.19\%). These results demonstrate that the performance gains stem intrinsically from the rich expressive capacity and information preservation of the latent space, rather than a mere increase in the number of agents. Notably, the superior performance of the \texttt{tree-3} configuration confirms that Interlat effectively supports parallel multi-agent reasoning, allowing downstream actors to synthesize heterogeneous latent perspectives without requiring pair-specific training.

\section{Conclusion}
In this work, we introduced \modelname, a paradigm that enables inter-agent communication entirely in latent space. Across experiments, our results show that directly transmitting and reasoning over latent states improves task performance and achieves substantially higher communication efficiency, even demonstrating compatibility across heterogeneous models.
Beyond full-length latent exchange, we show that latent communication can be aggressively compressed through latent-space reasoning, forming a compact, task-preserving representation that retains parallel hypotheses while discarding redundant structure.
Together, these findings suggest that communication need not be bound to language tokens, highlighting latent states as a viable, efficient, and generalizable medium for next-generation multi-agent systems.


\section{Limitations}
Our study has explored how inter-agent communication can be realized entirely in latent space and demonstrated its potential benefits in terms of performance and efficiency. However, several important limitations should be considered when interpreting our results and applying this paradigm more broadly.

First, our experiments mainly focus on a controlled two-agent setting, evaluated on one embodied interactive benchmark (ALFWorld) and one non-interactive symbolic reasoning benchmark (MATH). While these benchmarks jointly cover both interactive planning and single-turn reasoning scenarios, they do not yet fully capture the diversity of real-world multi-agent systems, such as settings with larger teams, dynamic role assignments, long-horizon collaboration, or richer environments involving tool use and external memory. Extending latent communication to larger-scale and more heterogeneous agent ecosystems is a natural and important direction for future work.

Second, our approach assumes access to internal model representations, specifically last-layer hidden states, in order to enable latent communication. This assumption may not hold for closed-source or API-only models, where hidden states are inaccessible. As a result, the current formulation of Interlat is primarily applicable to open or inspectable models, and extending latent communication to restricted-access settings is an open challenge.

Finally, latent communication trades human interpretability for efficiency and expressive capacity. Unlike language-based communication, latent messages are not directly human-readable, which complicates debugging, monitoring, and failure analysis in complex systems. While this work primarily focuses on task performance and efficiency and provides preliminary interpretability analysis via PCA (Figure~\ref{fig:pca}), developing principled tools to improve both the interpretability and controllability of latent communication remains an important direction for broader deployment.

Despite these limitations, we view this work as an initial yet concrete step toward understanding and enabling inter-agent communication beyond language. Our results provide empirical evidence that latent space communication can support effective coordination and can be aggressively compressed while preserving task utility, highlighting its potential as a complementary paradigm to language-based interaction in LLM-powered agent systems.

\section{Ethical Considerations}
No human participants, crowdsourcing, or personally identifiable information (PII) were involved in this research. All experiments were conducted within a simulated environment using standard dataset splits.

Our study focuses on inter-agent communication in latent space, utilizing the last hidden states and their compressed variants. A potential theoretical risk is that such latent communication could be exploited to circumvent language-based safety mechanisms. To mitigate this concern to the greatest extent possible, we neither trained on nor evaluated any harmful instructions, and no harmful actions occurred during our experiments. Furthermore, to promote transparency, we provide PCA-based visualizations of latent communication in Figure~\ref{fig:pca} the analyze internal probability distribution of compressed latent communications in Figure~\ref{fig:topk-6}, offering a clearer understanding of the information being transmitted.

\section{Acknowledgments}
This research was partially supported by National Key R\&D Program of China under Grant No. 2024YFF0907802,  Zhejiang Provincial Natural Science Foundation of China under Grant No. LD24F020011, and Alibaba Group through Alibaba Research Intern Program.

\bibliography{ref}

\newpage

\appendix

\newpage
\noindent\textbf{Appendix}

\noindent The supplementary information accompanying the main paper provides additional data, explanations, and details.

\section{LLM usage}
ChatGPT\footnote{https://chat.openai.com/} was used purely with the language of the paper during the writing process, including spell-checking and paraphrasing the authors' original content, without suggesting new content.
Any content generated with the assistant underwent meticulous manual review and subsequently received final approval from the authors.

\section{Compression Loss}\label{sec:compression}

\paragraph{Setup.}
After training an actor $M_{\theta}$ to consume latent communications, we freeze $M_{\theta}$ and train a reasoning model $M_{\phi}$ to \emph{produce} compact, information-dense latent communications of length $K$ that the frozen actor can still exploit.
For an input instance with supervised token indices $S$ (the teacher-forced window after the first user turn), let
\begin{equation}
    H^{\mathrm{gen}}_{1:K} \;=\; M_{\phi}(x)
    \quad\text{and}\quad
    H^{\mathrm{full}}_{1:L} \;=\; M_{\text{ins}}(x),
\end{equation}
denote respectively the \emph{generated} latent communication from the trainable reasoning model and the full-length latent communication extracted from a fixed instruction-tuned model $M_{\text{ins}}$.
A lightweight communication adapter $g(\cdot)$ (kept frozen) preprocesses the latent communication before concatenation with boundary tokens \texttt{<bop>}/\texttt{<eop>}. For brevity, we use $H_K \equiv H^{\mathrm{gen}}_{1:K}$ and $H_L \equiv H^{\mathrm{full}}_{1:L}$.

We define three actor-scored forward paths through the frozen actor $M_{\theta}$ given a prompt $x$:
(i) \textbf{Path A (generated latents)}: $E^{(A)}=[\,e(x),\,e(\texttt{<bop>}),\,g(H_K),\,e(\texttt{<eop>})\,]$;
(ii) \textbf{Path D (full-length latents)}: $E^{(D)}=[\,e(x),\,e(\texttt{<bop>}),\,g(H_L),\,e(\texttt{<eop>})\,]$;
(iii) \textbf{Path B (no latents)}: $E^{(B)}=[\,e(x)\,]$.
Let $\mathbf{z}^{(q)}_t$ be the frozen-actor logits at position $t\!\in\! S$ under path $q\!\in\!\{A,D,B\}$, and
\begin{equation}
p^{(q)}_t \;=\; \mathrm{softmax}\!\big(\mathbf{z}^{(q)}_t / T\big)
\end{equation}
be the corresponding token distributions with temperature $T\!\ge\!1$ used for distillation.
Unless stated otherwise, gradients do not flow into $M_{\theta}$ or $g(\cdot)$.

\paragraph{(1) Actor cross-entropy utility.}
We require the generated message to be \emph{useful} for the frozen actor:
\begin{equation}
\underbrace{\mathcal{L}_{\mathrm{task}}
\;=\;
\frac{1}{|S|} \sum_{t \in S}
\big(-\log p_{\theta}(y_t \mid C_t, H_K)\big)}_
\text{(computed under Path A).}
\end{equation}
This term enforces that the compressed latents $H_K$ still drive correct next-token predictions, directly penalizing information loss due to shortening ($K\!\ll\!L$).
It prevents degenerate “over-compression” that would be efficient but useless to the actor.
Practically, it anchors training on task utility, encouraging compression gain does not come at the cost of downstream performance.

\paragraph{(2) Uncertainty-weighted agreement.}
We further encourage \emph{behavioral agreement} between using full-length latent communication (Path D) and generated compressed latent communication (Path A), with per-token weights that reflect how much any latent reduces uncertainty relative to the no-latent baseline (Path B).
Let the entropies be
\begin{equation}
\underbrace{H^{(q)}(t) \;=\; -\sum_{v} p^{(q)}_t(v)\,\log p^{(q)}_t(v)}_{q\!\in\!\{A,D,B\}},.
\end{equation}
Define raw weights $w_t^\star=\max\!\big(H^{(B)}(t)-H^{(D)}(t),\,0\big)$ and optionally clip $w_t^\star$ to $[0,\tau]$ to suppress outliers. Normalize to unit mean:
\begin{equation}
w_t \;=\; \frac{w_t^\star}{\frac{1}{|S|}\sum_{u\in S} w_u^\star+\varepsilon}.
\end{equation}
The agreement term is a temperature-scaled KL:
\begin{equation}
\begin{split}
\mathcal{L}_{\mathrm{pref}}
&= \frac{1}{\sum_{t\in S} w_t}
   \sum_{t\in S} w_t \;
   T^2\,
   \mathrm{KL}\!\Big(p^{(D)}_t \,\big\|\, p^{(A)}_t\Big) \\
&= \frac{T^2}{\sum_{t\in S} w_t}
   \sum_{t\in S} w_t
   \sum_v p^{(D)}_t(v)
   \log\frac{p^{(D)}_t(v)}{p^{(A)}_t(v)}.
\end{split}
\end{equation}
By matching $p^{(A)}$ to $p^{(D)}$ where full latents actually reduce uncertainty (weights $w_t$), this term teaches $H_K$ to reproduce the \emph{informative} behavioral effects of $H_L$ while ignoring positions where latents are unhelpful. Unlike reconstruction-based or contrastive objectives, this formulation aligns compressed latents directly through the actor’s induced behavior, avoiding assumptions about latent invertibility or instance-level correspondence. This allows compressed communication to focus on functional equivalence rather than representational similarity. This is particularly important for reasoning latents, which are over-parameterized, temporally misaligned under compression, and lack a natural one-to-one mapping across steps. By aligning compressed and full communications through induced behavior, our formulation supports variable-length latent messages, enables abstraction across multiple reasoning steps, and yields more stable and transferable training signals.

\paragraph{(3) Latent direction alignment.}
To stabilize compression, we align the \emph{global direction} of actor-side latent features induced by generated vs.\ data latents.
Let $Z^{(q)}_{k} \in \mathbb{R}^{d_z}$ be the actor-side features (after $g(\cdot)$ and the actor’s input stack) at latent step $k$ under path $q\in\{A,D\}$.
When $H_L$ has length $L\!\neq\!K$, apply a fixed resampling operator $\rho_K$ (\eg uniform down/up-sampling) and write $Z^{(D)}_{1:K}=\rho_K\!\big(Z^{(D)}_{1:L}\big)$.
Define step-averaged directions $\bar{z}^{(q)}=\frac{1}{K}\sum_{k=1}^{K} Z^{(q)}_{k}$ and the cosine penalty
\begin{equation}
\begin{split}
\mathcal{L}_{\mathrm{geom}}
&= 1 - \cos\!\big(\bar{z}^{(A)},\,\bar{z}^{(D)}\big) \\
&= 1 - \frac{\langle \bar{z}^{(A)},\,\bar{z}^{(D)}\rangle}{\|\bar{z}^{(A)}\|_2\,\|\bar{z}^{(D)}\|_2}.
\end{split}
\end{equation}
This term preserves the \emph{geometry} of the actor-side representations, preventing the compressed latents from drifting to directions that the actor interprets differently.
Empirically, it improves stability and mitigates mode collapse when $K$ is small by retaining the global semantic orientation of $H_L$.

\paragraph{Overall objective.}
The compression objective for $M_{\phi}$ (with $M_{\theta}$ frozen) is

\begin{equation}
    \begin{split}
    \mathcal{L}_{\mathrm{compress}}
    &= \lambda_{\mathrm{task}}\;\mathcal{L}_{\mathrm{task}} \\
    &\quad + \lambda_{\mathrm{pref}}\;\mathcal{L}_{\mathrm{pref}} \\
    &\quad + \lambda_{\mathrm{geom}}\;\mathcal{L}_{\mathrm{geom}}.
    \end{split}
\end{equation}
In practice, all terms are computed over $t\!\in\!S$ with teacher forcing; gradients propagate only to $\phi$.

\subsection{Baselines and settings in \modelname.}\label{app:baseline}
We consider two external baselines, which do not rely on latent communication at all. All baselines are trained using the same base models (Qwen2.5-7B/0.5B-Base, LLaMA3.1-8B-Base) as \modelname, differing only in whether and how inter-agent communication is provided.

\begin{enumerate}
    \item \textbf{CoT (full).} 
    We use complete Chain-of-Thought (CoT) traces produced by a related instruction-tuned model (Qwen2.5-7B-Instruct, Qwen2.5-0.5B-Instruct, and LLaMA3.1-8B-Instruct) to perform full-parameter supervised fine-tuning. In inference, the model receives a complete CoT plan before generating answers. 
    
    \textbf{\emph{Rationale.}} This baseline serves as a strong upper bound for language-based communication: it evaluates whether latent communication can surpass explicit human-readable planning, and controls for the supervision quality provided by an instruction-tuned teacher.
    \item \textbf{No-CoT.} The language model is trained to produce the final answer directly, without receiving any plan from other agents.

    \textbf{\emph{Rationale.}} This baseline isolates the contribution of any communication signal. It tests whether inter-agent exchange, latent or linguistic, is necessary for solving multi-step tasks.
    
\end{enumerate}

In addition, we evaluate controlled \textbf{variants} of \modelname to diagnose what information is encoded in the latents.

\begin{enumerate}
    \item \textbf{Text.} Instead of latent communication, we feed the corresponding CoT plan (in language space) to the actor. 
    
    \textbf{\emph{Rationale.}} This variant keeps the interaction protocol unchanged while varying only the communication channel. It enables a direct comparison between language-space and latent-space communication under matched training conditions, disentangling architectural factors from representational ones.
    \item \textbf{No-Comm.} We remove any communication from the actor’s input. This variant quantifies the intrinsic benefit of communication in our framework and verifies that performance improvements do not arise solely from modifications to the underlying model parameters.
    \item \textbf{CrossTask.} We replace the current task’s latent communication with one sampled from a different task. 
    
    \textbf{\emph{Rationale.}} This variant examines whether the actor is genuinely interpreting task-specific latent content. A substantial degradation indicates reliance on meaningful information encoded in the latents, rather than superficial distributional shortcuts.
    \item \textbf{Noised.} We add perturbations to the latent communication $H$: 
    (a) \textbf{CovNoise-0.5$\times$/1.0$\times$}: covariance-shaped noise $\varepsilon_t \sim \mathcal{N}(0,\hat{\Sigma})$ with optional strength $\lambda \in \{0.5,1.0\}$, where $\hat{\Sigma}$ is the sample covariance of the original $H$; 
    (b) \textbf{WhiteNoise}: a control drawn from $\mathcal{N}(0,I)$ with the same length. 
    
    \textbf{\emph{Rationale.}} Noise-based perturbations interrogate the robustness and locality of latent-space semantics. Covariance-shaped noise preserves global second-order structure, whereas white noise does not, allowing us to assess whether the actor relies on fine-grained geometric relations within true latent trajectories.
    \item \textbf{CovGauss.} We replace the entire $H$ with \iid samples $H_t \sim \mathcal{N}(0,\hat{\Sigma})$ (0$\mu$) and report a robustness check with $\mathcal{N}(\hat{\mu},\hat{\Sigma})$ ($\mu$). These preserve first-second order moments while removing higher-order structure and temporal alignment. 
    
    \textbf{\emph{Rationale.}} This variant preserves the mean and covariance of the original latent distribution while discarding all higher-order statistics and temporal correlations. It tests whether latent communication conveys information beyond global moments (first-order and second-order moments), \eg structured reasoning paths or non-Gaussian manifold geometry.
    \item \textbf{RandomRot.} We apply a structure-preserving but information-scrambling transform $H'=\hat{\mu}+(H-\hat{\mu})\,\hat{\Sigma}^{-1/2}\,Q\,\hat{\Sigma}^{1/2}$, where $Q$ is a Haar-random orthogonal matrix~\citep{mezzadri2006generate}. 
    
    \textbf{\emph{Rationale.}} This preserves the mean/covariance exactly while disrupting higher-order structure. Random rotation strictly preserves the first two moments of the latent distribution while scrambling its geometric orientation and higher-order structure. This constitutes a strong diagnostic of whether the actor depends on directional semantics or sequential organization within the latent manifold, rather than mere distributional similarity.
    \item \textbf{Cross-Family.}
    We evaluate \modelname under a cross-model-family setting, where the sender and actor belong to different pretrained model families.
    Specifically, latent communications are generated by a sender model from one family (In this work, we use Qwen2.5-7B-Instruct) and consumed by an actor model from another family (LLaMA3.1-8B-Base), without sharing parameters or tokenizer vocabularies.
    
    \textbf{\emph{Rationale.}} This setting tests whether latent communication encodes task-relevant information in a model-agnostic manner, rather than exploiting family-specific activation conventions or implicit alignment.
\end{enumerate}

\section{Benchmark}\label{app:why-alfworld}
\subsection{Alfworld}
Alfworld~\cite{shridhar2020alfworld} is a text-only benchmark that simulates embodied household tasks while keeping interaction purely in natural language. Agents observe textual descriptions of the scene and issue high-level commands from a constrained action set (\eg go to, open, close, take, put, toggle on/off, heat, cool, examine). Tasks are long-horizon and compositional, requiring perception, planning, and execution over multiple steps under partial observability. The benchmark provides official train/seen/unseen splits and a standard success metric under a fixed step budget (\eg 20 steps in our setup), enabling systematic and reproducible evaluation of sequential decision-making.

\begin{figure}
    \centering
    \includegraphics[width=1\linewidth]{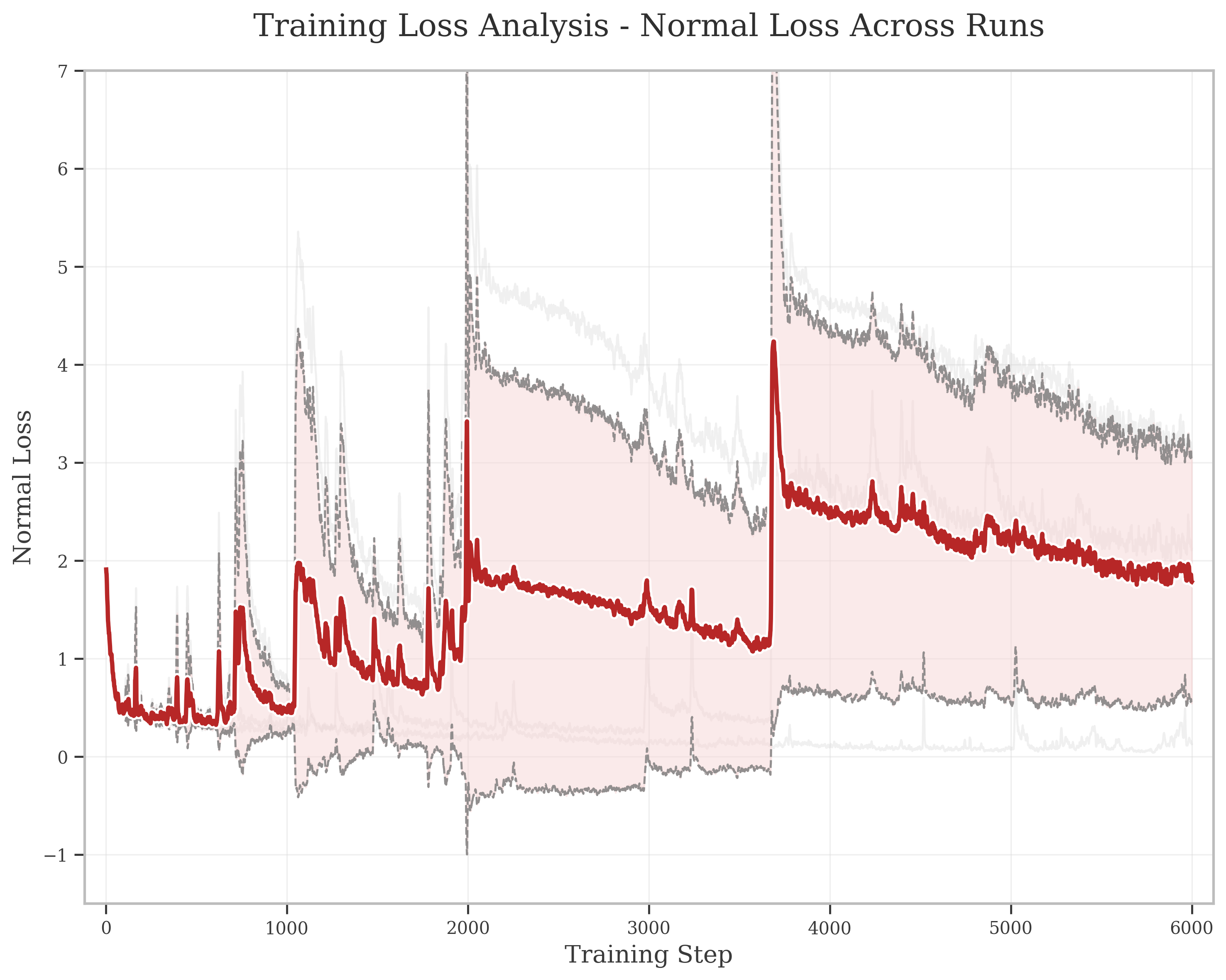}
    \caption{Training dynamics of the cross-entropy loss when curriculum learning is removed, illustrating highly unstable optimization behavior.}
    \label{fig:ablation_curriculum}
\end{figure}

\section{Ablations and Step Analysis}\label{app:ablation}
We present ablation studies for both the actor and reasoning models, reporting the average number of steps for successful trials versus all trials (success/all). 

\paragraph{Effect of curriculum learning.}
For the actor model, removing curriculum learning forces the agent to interpret latent communications from scratch. As shown in Figure~\ref{fig:ablation_curriculum}, this leads to highly unstable training dynamics and substantially degraded latent comprehension, preventing the model from consistently leveraging the communicated information.

\paragraph{Step count versus performance.}
Table~\ref{tab:ablation_full} reveals a nuanced but systematic relationship between step count and task performance. On seen tasks, ablating key components results in a lower overall success rate. Although these ablated models often take fewer steps on the trials they complete, their high failure rate indicates an inability to reliably interpret latent communication and solve tasks. In contrast, the full model achieves both higher success rates and longer trajectories, suggesting that additional steps correspond to productive exploration rather than inefficiency.

On unseen tasks, several ablations (\eg removing curriculum learning or the communication adapter) exhibit the opposite pattern: the agent takes more steps while achieving a lower success rate. This demonstrates that longer trajectories alone do not imply effective exploration. Without these critical components, the policy exhibits unstructured search behavior that fails to form coherent task-solving strategies. Together, these observations underscore the importance of evaluating step count jointly with success rate, and support our central claim that information-rich latent communication enables structured and effective exploration rather than random wandering.

\begin{table*}[htbp]
\centering
\setlength{\tabcolsep}{6pt} 
\renewcommand{\arraystretch}{1.05} 
\begin{tabular}{lrrrr}
\toprule
Method & Seen & Steps & Unseen & Steps \\
\midrule[\cmidrulewidth]  
\multicolumn{5}{l}{\textbf{Actor model}}\\
\cmidrule(lr){1-5}
Ours Full & $\textbf{70.48}_{\pm 1.01}$ & 9.41/12.54 & $\textbf{65.42}_{\pm 0.87}$ & 9.86/13.37 \\
\cmidrule(lr){1-5} 
$\text{w/o}$ curri & $33.10_{\pm 2.97}$ & 9.07/16.38 & $20.65_{\pm 2.15}$ & 10.47/18.03 \\
$\text{w/o}$ $\mathcal{L}_\text{sep}$ & $\underline{58.81}_{\pm 1.41}$ & 8.07/12.98 & $\underline{60.70}_{\pm 5.50}$ & 9.64/13.71 \\
$\text{w/o}$ $\mathcal{L}_\text{align}$ & $56.90_{\pm 1.41}$ & 8.16/13.26 & $53.98_{\pm 3.35}$ & 9.56/14.36 \\
$\text{w/o}$ adapter & $4.05_{\pm 1.70}$ & 9.32/19.57 & $4.48_{\pm 1.31}$  & 10.53/19.58 \\
\midrule[\cmidrulewidth]  
\multicolumn{5}{l}{\textbf{Reasoning model}}\\
\cmidrule(lr){1-5}
Ours Full & $\textbf{68.10}_{\pm 1.93}$ & 9.21/12.65 & $\underline{62.94}_{\pm 2.03}$ & 9.88/13.63 \\
\cmidrule(lr){1-5}
$\text{w/o}$ $\mathcal{L}_{\mathrm{task}}$ & $\underline{65.71}_{\pm 1.43}$ & 8.86/12.68 & $\textbf{63.18}_{\pm 3.47}$ & 9.68/13.48 \\
$\text{w/o}$ $\mathcal{L}_\text{pref}$ & $64.76_{\pm 2.97}$ & 8.92/12.82 & $60.20_{\pm 3.13}$ & 9.68/13.79 \\
$\text{w/o}$ $\mathcal{L}_\text{geom}$ & $64.05_{\pm 3.55}$ & 8.71/12.77 & $59.45_{\pm 3.01}$ & 9.88/13.98 \\
\bottomrule
\end{tabular}
\caption{Ablation of training components. “Ours Full” uses all components.}
\label{tab:ablation_full}
\end{table*}

\section{Compression Result}\label{app:compression}
\begin{figure}[t]
    \centering 
    \subfigure[Training-free sweep over retained ratio $R$]{ 
        \label{fig:compression-left} 
        \includegraphics[width=2in]{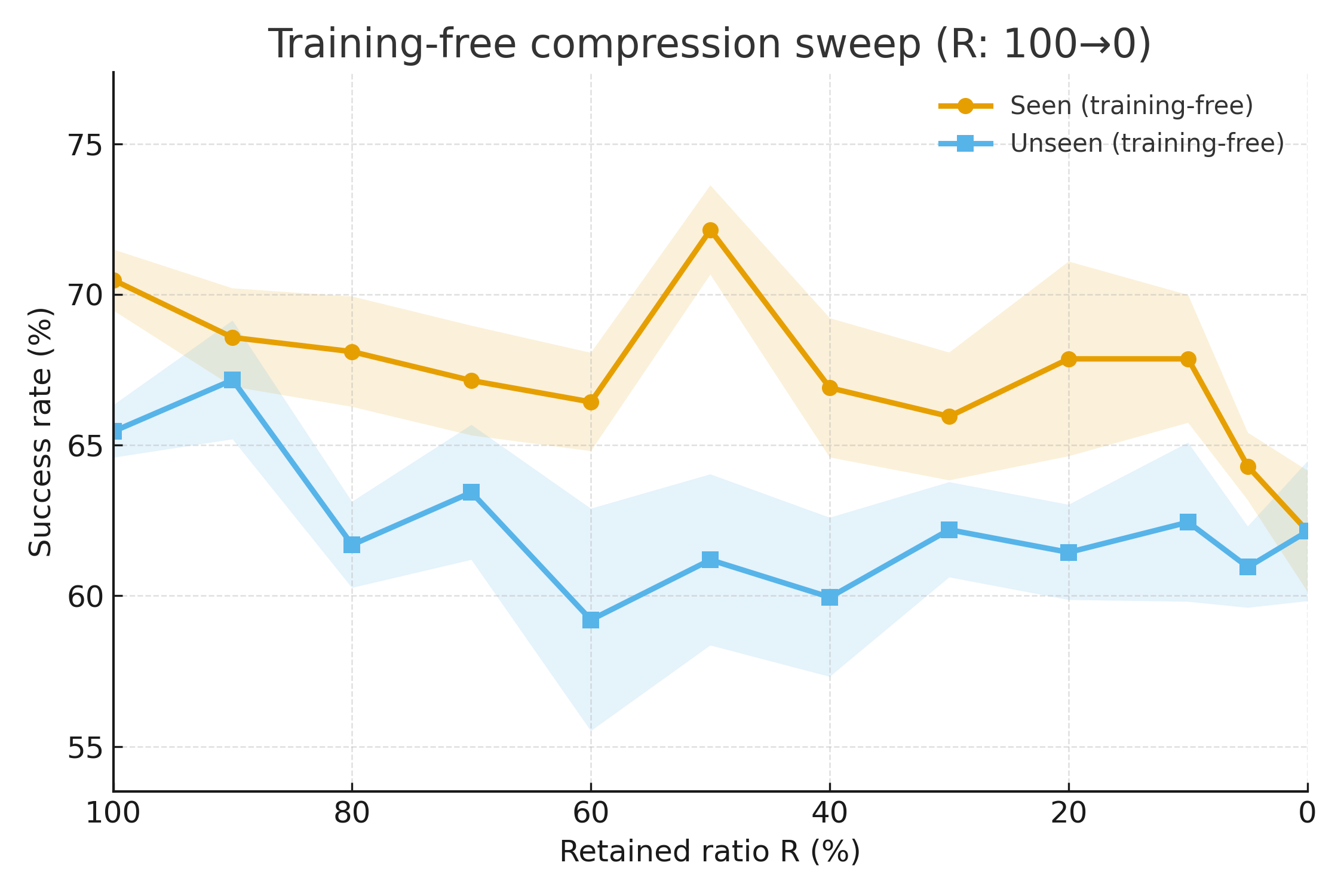}} 
    \hspace{1in} 
    \subfigure[Latent length $L$ vs. performance (untrained \& trained)]{ 
        \label{fig:compression-right} 
        \includegraphics[width=2in]{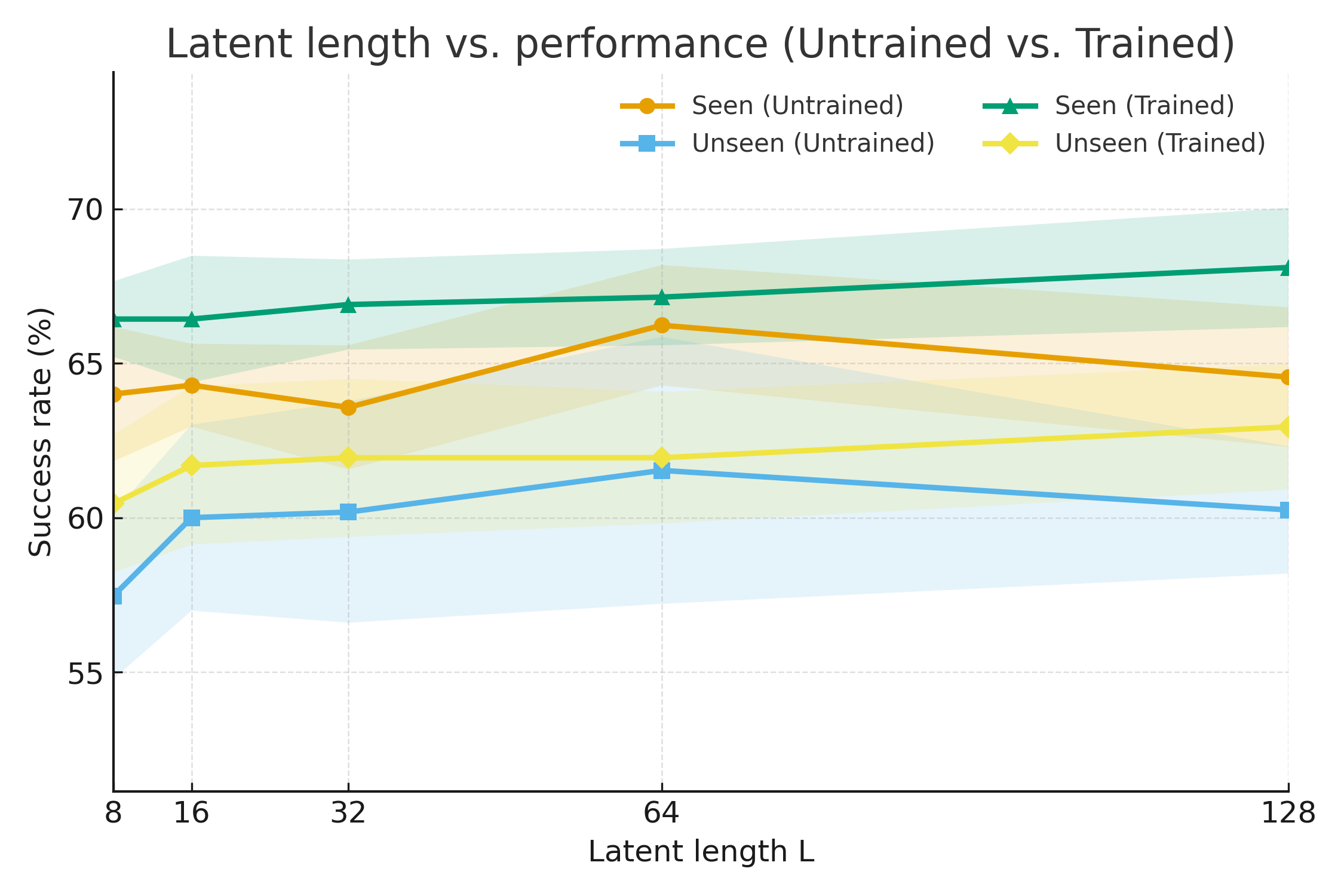}} 
    \caption{Result of compression on seen and unseen tasks. \textbf{Left:} Success rate under training-free compression with different retained ratios $R$. \textbf{Right:} Performance of untrained and trained models across latent lengths $L$} 
    \label{fig:compression} 
\end{figure}
In this section, we provide more detailed results on compression with average steps as success/all across tasks in Table~\ref{tab:compression_full_LLaMA} (LLaMA3.1-8B-Base) and Table~\ref{tab:compression_full_qwen} (Qwen2.5-7B-Base) and corresponding performance trend in Figure~\ref{fig:compression}. Latency is measured on the same machine and decoding policy (if needed) across rows~\footnote{For the untrained reasoning model, we use the standard
\texttt{generate} API from Hugging Face \texttt{transformers}; see
\url{https://github.com/huggingface/transformers}.}.

\begin{table*}[t]
\centering
\begin{tabular}{llllll}
\toprule[1.5pt]
\multicolumn{6}{c}{\textbf{LLaMA3.1-8B}} \\
\toprule[1.5pt]
Ratio & Seen & Steps & Unseen & Steps & Time \\
\toprule[1pt]
Untrained \\
\toprule[0.25pt]
Full   & $\textbf{70.71}_{\pm 1.04}$ & 8.02/12.58 & $\textbf{70.90}_{\pm 1.21}$ & 8.21/12.96 & 10.20s \\
90\%   & $60.09_{\pm 2.94}$ & 8.12/12.86 & $60.45_{\pm 1.58}$ & 9.51/13.66 & - \\
80\%   & $\uline{65.27}_{\pm 1.66}$ & 8.23/12.32 & $\uline{63.43}_{\pm 4.84}$ & 9.05/13.06 & - \\
70\%   & $60.62_{\pm 2.43}$ & 8.14/12.81 & $61.79_{\pm 2.81}$ & 10.13/13.90 & - \\
60\%   & $59.82_{\pm 2.55}$ & 7.90/12.76 & $63.44_{\pm 2.79}$ & 9.39/13.27 & - \\
50\%   & $63.12_{\pm 3.58}$ & 8.02/12.44 & $60.07_{\pm 1.43}$ & 9.85/13.90 & - \\
40\%   & $65.27_{\pm 1.08}$ & 8.06/12.21 & $59.70_{\pm 1.83}$ & 9.67/13.83 & - \\
30\%   & $61.25_{\pm 2.61}$ & 8.20/12.77 & $59.51_{\pm 2.39}$ & 10.13/14.13 & - \\
20\%   & $61.79_{\pm 3.10}$ & 8.41/12.84 & $57.84_{\pm 1.58}$ & 9.48/13.92 & - \\
10\%   & $64.91_{\pm 1.23}$ & 8.60/12.60 & $60.45_{\pm 4.22}$ & 9.48/13.64 & - \\
5\%    & $63.68_{\pm 2.57}$ & 8.42/12.62 & $60.95_{\pm 1.35}$ & 9.90/13.84 & - \\
0\%    & $63.57_{\pm 2.44}$ & 8.35/12.59 & $58.40_{\pm 2.76}$ & 9.47/13.85 & - \\
128L   & $61.59_{\pm 2.34}$ & 8.24/12.75 & $62.39_{\pm 2.82}$ & 10.01/13.77 & 4.00s \\
64L    & $62.54_{\pm 2.90}$ & 8.38/12.73 & $60.52_{\pm 3.08}$ & 9.86/13.86 & 2.10s \\
32L    & $62.30_{\pm 2.08}$ & 8.33/12.73 & $57.46_{\pm 2.28}$ & 9.98/14.24 & 1.20s \\
16L    & $63.89_{\pm 2.70}$ & 8.53/12.67 & $59.61_{\pm 3.53}$ & 9.50/13.74 & 0.70s \\
8L     & $61.79_{\pm 2.44}$ & 8.09/12.64 & $58.77_{\pm 2.76}$ & 9.99/14.12 & 0.45s \\
\toprule[1pt]
Trained \\
\toprule[0.25pt]
128L & $66.46_{\pm 1.98}$ & 8.18/12.54 & $66.35_{\pm 1.86}$ & 8.96/13.12 & 2.80s \\
64L  & $66.21_{\pm 1.72}$ & 8.12/12.58 & $65.42_{\pm 1.94}$ & 9.02/13.20 & 1.40s \\
32L  & $65.45_{\pm 1.63}$ & 8.08/12.60 & $65.01_{\pm 1.88}$ & 9.08/13.28 & 0.72s \\
16L  & $64.41_{\pm 1.95}$ & 8.10/12.64 & $65.20_{\pm 1.76}$ & 9.12/13.34 & 0.39s \\
8L   & $64.32_{\pm 1.84}$ & 8.14/12.66 & $64.89_{\pm 1.69}$ & 9.18/13.40 & 0.24s \\
\toprule[1.5pt]
\end{tabular}
\caption{Complete compression results with seen/unseen accuracy, steps, and latency across tasks on LLaMA models.}
\label{tab:compression_full_LLaMA}
\end{table*}

\begin{table*}[t]
\centering
\begin{tabular}{llllll}
\toprule[1.5pt]
\multicolumn{6}{c}{\textbf{Qwen2.5-7B}} \\
\toprule[1.5pt]
Ratio & Seen & Steps & Unseen & Steps & Time \\
\toprule[1pt]
Untrained \\
\toprule[0.25pt]
Full   & $\underline{70.48}_{\pm 1.01}$ & 9.41/12.54 & $\underline{65.42}_{\pm 0.87}$ & 9.86/13.37 & 9.19s \\
90\%    & $68.57_{\pm 1.63}$ & 8.77/12.30 & $\textbf{67.16}_{\pm 1.97}$ & 9.27/12.79 & - \\
80\%    & $68.10_{\pm 1.83}$ & 8.56/12.21 & $61.69_{\pm 1.43}$ & 9.10/13.28 & - \\
70\%   & $67.14_{\pm 1.82}$ & 8.68/12.40 & $63.43_{\pm 2.24}$ & 9.42/13.29 & - \\
60\%   & $66.43_{\pm 1.63}$ & 8.52/12.37 & $59.20_{\pm 3.69}$ & 9.90/14.02 & - \\
50\%   & $\textbf{72.14}_{\pm 1.48}$ & 9.03/12.09 & $61.19_{\pm 2.84}$ & 9.37/13.50 & - \\
40\%   & $66.90_{\pm 2.31}$ & 8.88/12.56 & $59.95_{\pm 2.64}$ & 9.52/13.72 & - \\
30\%   & $65.95_{\pm 2.12}$ & 8.80/12.61 & $62.19_{\pm 1.58}$ & 10.11/13.85 & - \\
20\%   & $67.86_{\pm 3.23}$ & 8.97/12.52 & $61.44_{\pm 1.58}$ & 9.98/13.84 & - \\
10\%   & $67.86_{\pm 2.12}$ & 8.76/12.37 & $62.44_{\pm 2.64}$ & 9.72/13.58 & - \\
5\%    & $64.52_{\pm 1.12}$ & 9.19/13.02 & $60.95_{\pm 1.35}$ & 9.90/13.84 & - \\
0\%    & $62.14_{\pm 2.01}$ & 10.19/13.90 & $62.19_{\pm 2.32}$ & 10.23/13.92 & - \\
128L   & $64.52_{\pm 2.26}$ & 8.68/12.70 & $60.20_{\pm 2.06}$ & 9.69/13.79 & 3.55s \\
64L    & $66.19_{\pm 1.95}$ & 8.76/12.56 & $61.44_{\pm 4.32}$ & 9.85/13.76 & 1.83s \\
32L    & $63.57_{\pm 2.01}$ & 8.66/12.79 & $60.20_{\pm 3.58}$ & 9.87/13.90 & 1.03s \\
16L    & $64.29_{\pm 1.34}$ & 8.64/12.70 & $59.95_{\pm 3.01}$ & 10.07/14.05 & 0.62s \\
8L     & $64.05_{\pm 2.18}$ & 8.80/12.83 & $57.46_{\pm 2.69}$ & 10.29/14.42 & 0.39s \\
\toprule[1pt]
Trained \\
\toprule[0.25pt]
128L & $68.10_{\pm 1.93}$ & 9.21/12.65 & $62.94_{\pm 2.03}$ & 9.88/13.63 & 2.25s \\
64L  & $67.14_{\pm 1.56}$ & 9.15/12.72 & $61.94_{\pm 2.13}$ & 9.92/13.76 & 1.16s \\
32L  & $66.90_{\pm 1.46}$ & 9.02/12.65 & $61.94_{\pm 2.56}$ & 9.96/13.78 & 0.60s \\
16L  & $66.43_{\pm 2.05}$ & 9.08/12.75 & $61.69_{\pm 2.56}$ & 9.98/13.82 & 0.33s \\
8L   & $66.43_{\pm 1.22}$ & 9.11/12.77 & $60.45_{\pm 2.23}$ & 9.90/13.89 & 0.20s \\
\toprule[1.5pt]
\end{tabular}
\caption{Complete compression results with seen/unseen accuracy, steps, and latency across tasks on Qwen models.} 
\label{tab:compression_full_qwen}
\end{table*}

\begin{figure*}[htbp]
  \centering
  \includegraphics[width=\linewidth]{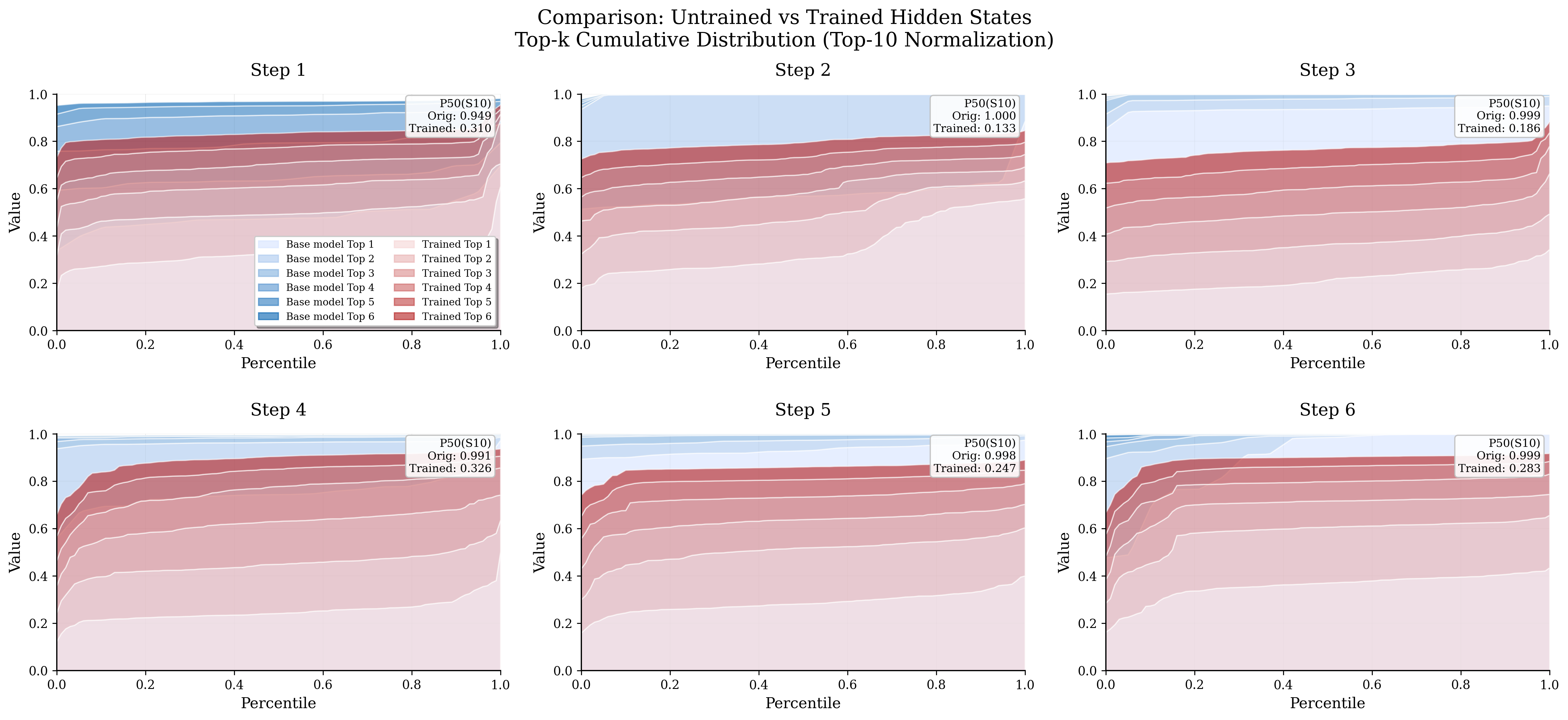}
  \caption{Parallelism in latent communication over the first six steps.
    Red indicates latents from the trained model, and blue indicates latents from the untrained base model.
    The trained latents preserve stable vertical gaps between successive Top-$k$ bands and achieve a markedly lower $P_{50}(S_{10})$,
    evidencing persistent parallelism, whereas the untrained base model's latents progressively collapse toward Top-1.}
  \label{fig:topk-6_base}
\end{figure*}

\begin{figure*}[htbp]
  \centering
  \includegraphics[width=\linewidth]{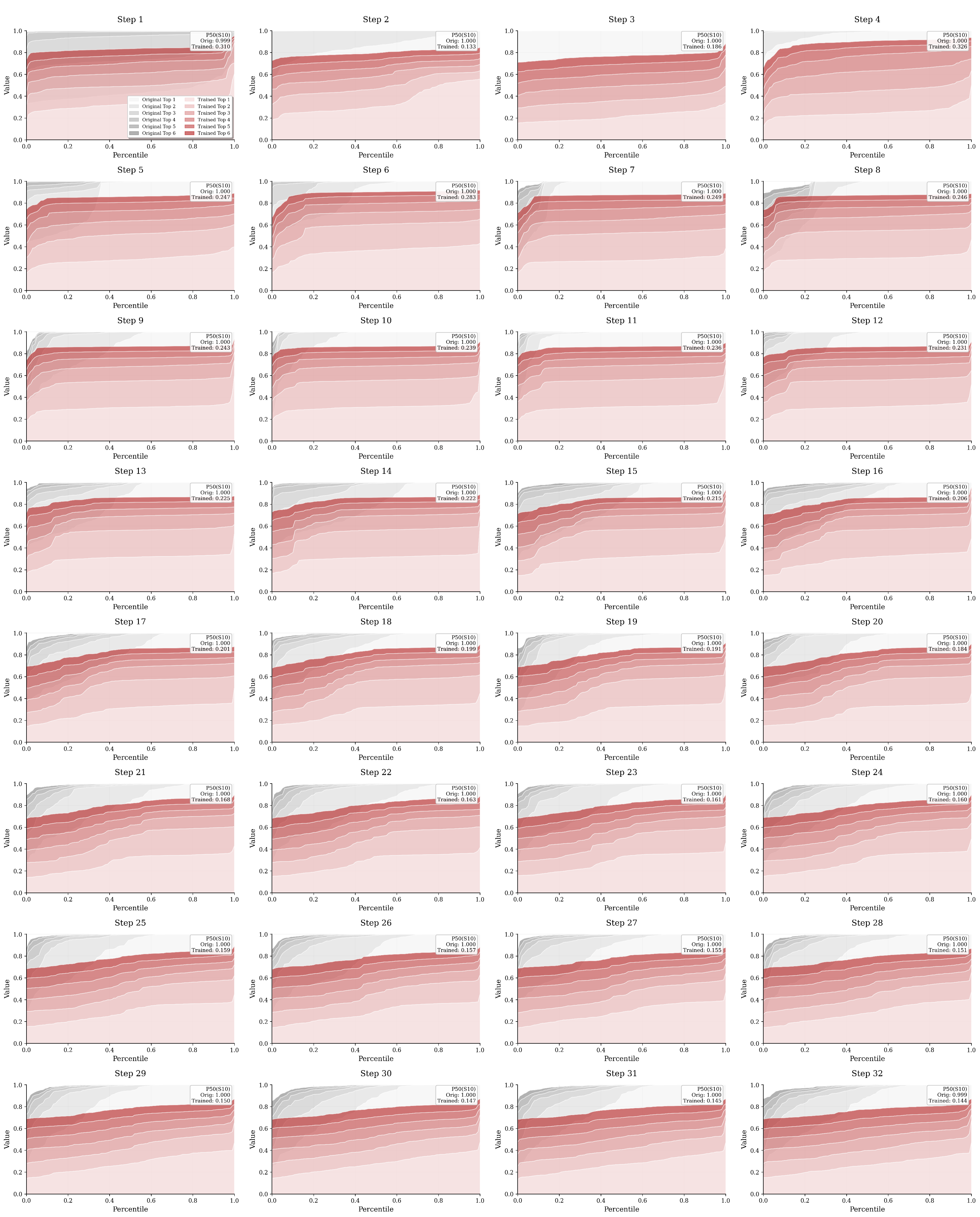}
    \caption{\textbf{Extended analysis (32 steps).}
  Same construction as Fig.~\ref{fig:topk-6}, now for steps $1\text{--}32$. Persistent separation among successive Top-$k$ bands and consistently lower $\mathrm{P50}(S_{10})$ values indicate that the trained latents maintain broad, plausible reasoning branches across the entire sequence, despite compression.}
  \label{fig:topk-32}
\end{figure*}

\section{Latent Parallelism Analysis}\label{sec:parallelism}
We first compared the latent communications produced by our trained reasoning model with those from an off-the-shelf Qwen2.5-7B-Instruct model in the compression-effectiveness analysis (see the Experiments section). Because our reasoning model is initialized from Qwen2.5-7B-Base, we additionally compare it with this base model, which has not been trained for generating compressed latent communication, in Figure~\ref{fig:topk-6_base}. The findings are consistent with the earlier comparison: the trained model maintains stable vertical gaps between successive Top-$k$ curves across steps and exhibits a substantially lower $P_{50}(S_{10})$, whereas the base model shows a clear convergence toward Top-1.

We further extend the parallelism analysis to a deeper horizon of 32 steps. As shown in Figure~\ref{fig:topk-32}, the trained model exhibits stable vertical gaps between successive Top-$k$ curves throughout these steps. This extended analysis further verifies that the trained latent representations preserve a broader set of plausible reasoning paths by sustaining a more balanced probability distribution rather than prematurely collapsing to a Top-1 hypothesis.

\begin{figure}[htbp]
  \centering
  \includegraphics[width=\linewidth]{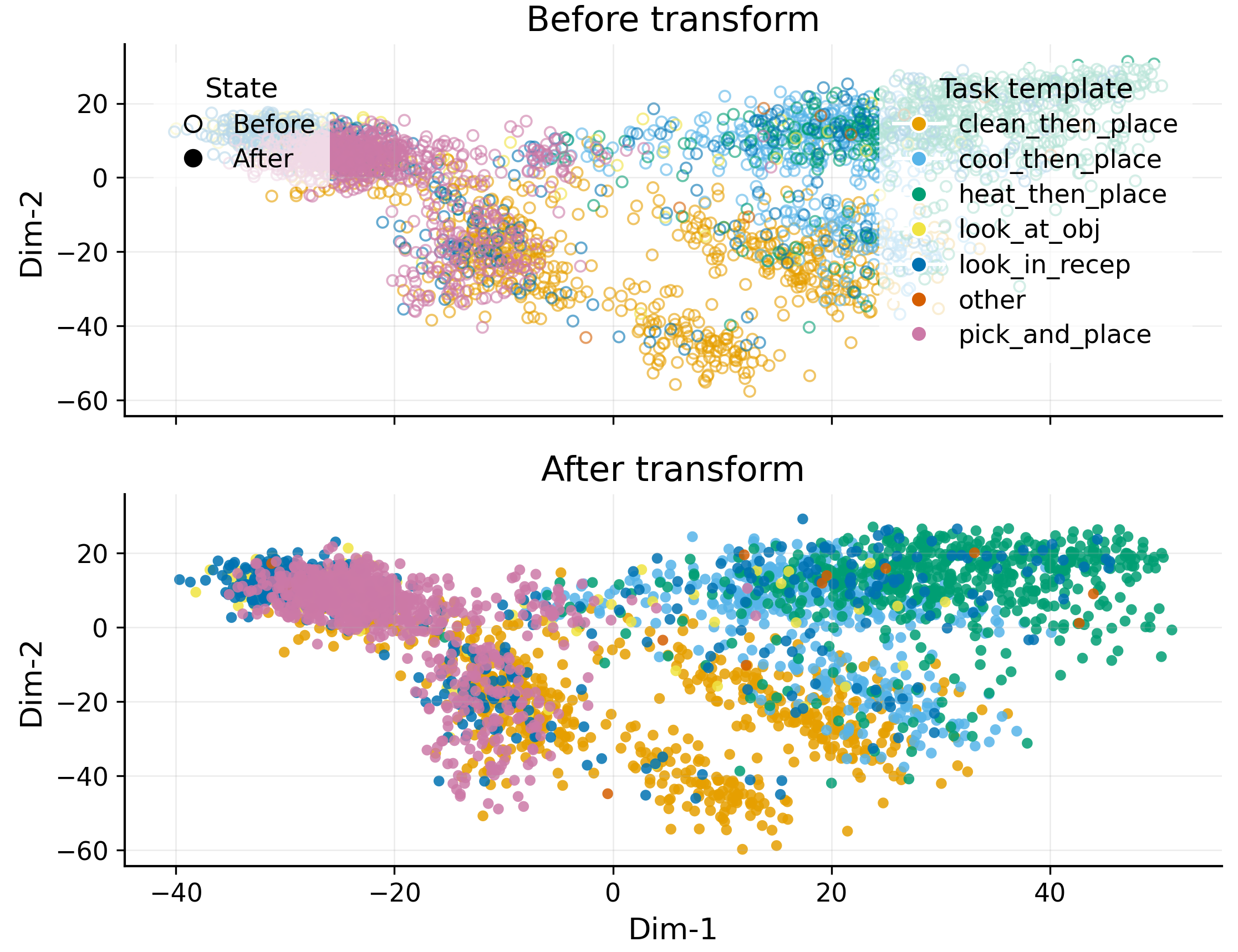}
  \caption{PCA visualization of latent communications grouped by ALFWorld task templates ($N=3{,}119$). Each point corresponds to the mean-pooled last-layer hidden state of the reasoning agent’s plan for a specific task instance. Colors indicate six core task templates: \texttt{pick\_and\_place}, \texttt{pick\_clean\_then\_place}, \texttt{pick\_heat\_then\_place}, \texttt{pick\_cool\_then\_place}, \texttt{look\_in\_recep}, and \texttt{look\_at\_obj}. The emergence of distinct clusters suggests that Interlat’s latent communication captures task-specific semantic structure, enabling the actor agent to differentiate diverse reasoning paradigms without relying on natural language.}
  \label{fig:cluster}
\end{figure}

\begin{figure}[htbp]
  \centering
  \includegraphics[width=\linewidth]{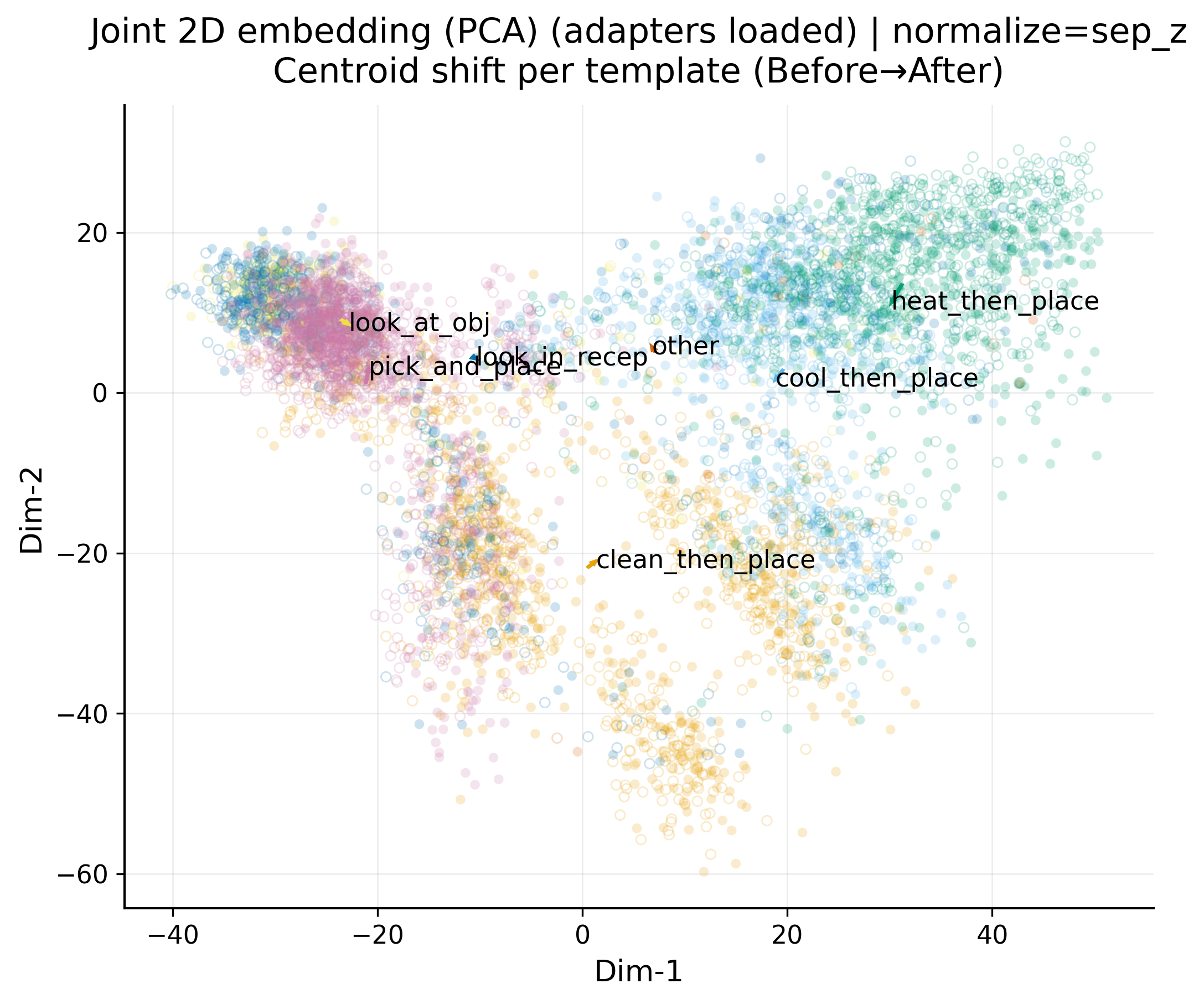}
 \caption{
    Joint PCA visualization of latent communications grouped by ALFWorld task templates.
    Each point represents the mean-pooled last-layer hidden state of the reasoning agent for a single task instance ($N=3{,}119$).
    Colors indicate different task templates.
    Solid points correspond to latent representations after applying the communication adapter, while hollow points denote the original latents before transformation.
    Arrows depict the centroid shift for each template from before to after.
    Although the centroid shifts are moderate in magnitude, their directions are consistently template-dependent, indicating structured semantic reorganization rather than global rescaling of the latent space.
}
  \label{fig:centroid_shift}
\end{figure}

\begin{table*}[t]
\centering
\small
\begin{tabular}{lrrrr}
\toprule
Task template & \#Samples & $L$ & $R$ (median) & $\rho = L/R$ \\
\midrule
pick\_and\_place      & 848 & 1.742 & 8.347  & 0.209 \\
heat\_then\_place     & 578 & 2.403 & 11.506 & 0.209 \\
cool\_then\_place     & 429 & 1.329 & 13.111 & 0.101 \\
look\_at\_obj         & 187 & 0.794 & 8.498  & 0.093 \\
clean\_then\_place    & 632 & 0.989 & 17.839 & 0.055 \\
look\_in\_recep       & 431 & 0.250 & 23.393 & 0.011 \\
other                 & 14  & 0.581 & 32.346 & 0.018 \\
\bottomrule
\end{tabular}
\caption{Quantitative centroid shift analysis for each task template.}
\label{tab:centroid_shift}
\end{table*}

\section{Qualitative Analysis of Latent Communication via PCA}
\label{app:pca}

To qualitatively examine the semantic structure encoded in latent communications, we perform Principal Component Analysis (PCA) on 3,119 samples from the ALFWorld training set. Each sample corresponds to the mean-pooled last-layer hidden state generated by the reasoning agent for a specific task instance. Tasks are grouped according to the official ALFWorld task templates, which define six core reasoning patterns: \texttt{pick\_and\_place}, \texttt{pick\_clean\_then\_place}, \texttt{pick\_heat\_then\_place}, \texttt{pick\_cool\_then\_place}, \texttt{look\_in\_recep}, and \texttt{look\_at\_obj}. Figure~\ref{fig:cluster} visualizes the projection of latent communications onto the first two principal components, which capture the dominant axes of variance across task instances.

The resulting PCA visualization reveals clear task-dependent organization in the latent space. Action-centric templates such as \texttt{pick\_and\_place} form a dense central cluster, while templates involving additional procedural constraints—such as thermal manipulation in \texttt{pick\_heat\_then\_place} and \texttt{pick\_cool\_then\_place}—occupy adjacent yet separable regions. Perception-oriented tasks (\texttt{look\_in\_recep} and \texttt{look\_at\_obj}), although less frequent, also exhibit localized concentrations distinct from execution-heavy templates. As PCA preserves global variance structure rather than emphasizing local neighborhoods, this separation indicates that task-specific semantics are encoded in the dominant latent dimensions, rather than arising from projection artifacts. Moreover, intra-cluster dispersion (e.g., within \texttt{pick\_and\_place}) suggests that latent representations retain fine-grained variations across task instances, rather than collapsing to a single prototype per template.

Beyond static clustering, we further analyze how these task-level structures are affected by the latent communication adapter. As shown in Figure~\ref{fig:centroid_shift} and quantified in Table~\ref{tab:centroid_shift}, different task templates exhibit consistent but template-dependent centroid shifts before and after transformation. While the absolute magnitudes of these shifts are moderate relative to within-template dispersion, their directions are highly structured, indicating selective reorganization rather than global rescaling of the latent space. Execution-heavy templates such as \texttt{pick\_and\_place} and \texttt{pick\_heat\_then\_place} undergo larger relative shifts, whereas observation-oriented templates remain more stable. Together, these results suggest that \modelname{} preserves the overall geometry of latent communications while inducing task-aware semantic alignment, enabling the actor agent to differentiate diverse reasoning paradigms without relying on explicit natural language communication.

\section{Training Template}\label{app:template}
We present an example in Figure~\ref{fig:conversations} to illustrate how agents explore and solve tasks in Alfworld. After perceiving the environment, the agent executes an action, receives feedback from the environment, and then proceeds to the next step until the goal is accomplished. Figure~\ref{fig:train_template} shows an example of how training data is structured for the actor agent. We append either the latent communication or the natural-language plan after the instruction to facilitate inter-agent communication.
\begin{templatebox}[Conversations]\label{fig:conversations}
  \envbubble{You are in the middle of a room. Looking quickly around you, you see a bathtubbasin 1, a cabinet 2, a cabinet 1, a countertop 1, a garbagecan 1, a handtowelholder 1, a sinkbasin 1, a toilet 1, a toiletpaperhanger 1, and a towelholder 1.\\
  \textbf{Task:} put a toiletpaper in toiletpaperhanger.}
  \vspace{6pt}
  \agentbubble{\textbf{Thought}: I need to place a toilet paper roll on the toilet paper hanger. First, I should check if there's already a roll on the hanger or if it's empty.\\
  \textbf{Action}: go to toiletpaperhanger 1}
  \vspace{6pt}
  \envbubble{\textbf{Observation}: On the toiletpaperhanger 1, you see nothing.}
  \vspace{6pt}
  \agentbubble{\textbf{Thought}: The toilet paper hanger is empty, so I need to find a new roll of toilet paper to put on it.\\
  \textbf{Action}: go to toilet 1}
  \vspace{6pt}
  \envbubble{\textbf{Observation}: On the toilet 1, you see a soapbottle 1, and a toiletpaper 1.}
  \vspace{6pt}
  \agentbubble{\textbf{Thought}: I've found a toilet paper roll on the toilet. I should pick it up so I can place it on the hanger.\\
  \textbf{Action}: take toiletpaper 1 from toilet 1}
  \vspace{6pt}
  \envbubble{\textbf{Observation}: You pick up the toiletpaper 1 from the toilet 1.}
  \vspace{6pt}
  \agentbubble{\textbf{Thought}: Now that I have the toilet paper roll, I need to return to the toilet paper hanger to install it.\\
  \textbf{Action}: go to toiletpaperhanger 1}
  \vspace{6pt}
  \envbubble{\textbf{Observation}: On the toiletpaperhanger 1, you see nothing.}
  \vspace{6pt}
  \agentbubble{\textbf{Thought}: I'm back at the toilet paper hanger with the roll in hand. It's time to put the toilet paper on the hanger.\\
  \textbf{Action}: put toiletpaper 1 in/on toiletpaperhanger 1}
  
\end{templatebox}

\begin{templatebox}[Training Template for Actor Agent]\label{fig:train_template}
\sys\\ You are a helpful assistant.\ \imend \\
\user\\ Interact with a household to solve a task. Imagine you are an intelligent agent in a household environment and your target is to perform actions to complete the task goal. At the beginning of your interactions, you will be given a detailed description of the current environment and your goal to accomplish. \\For each of your turns, you will be given the observation of the last turn. You should first think about the current condition and plan for your future actions, and then output your action in this turn. Your output must strictly follow this format:"Thought: your thoughts.\\Action: your next action".\\The available actions are:\\1. go to {recep}\\2. task {obj} from {recep}\\3. put {obj} in/on {recep}\\4. open {recep}\\5. close {recep}\\6. toggle {obj} {recep}\\7. clean {obj} with {recep}\\8. heat {obj} with {recep}\\9. cool {obj} with {recep}\\where {obj} and {recep} correspond to objects and receptacles.\\After your each turn, the environment will give you immediate feedback based on which you plan your next few steps. if the envrionment output "Nothing happened", that means the previous action is invalid and you should try more options.\\Your response should use the following format:\\Thought: <your thoughts>\\Action: <your next action>\\You are in the middle of a room. Looking quickly around you, you see a armchair 2, a armchair 1, a coffeetable 2, a coffeetable 1, a diningtable 1, a garbagecan 1, a sidetable 2, a sidetable 1, and a sofa 1.\\Your task is to: find two newspaper and put them in armchair.\\Now, you are given a step-by-step plan to complete this task as follow: \\
\texttt{<bop>}\\
\uline{Insert a latent communication or textual plan for inter-agent communication.} \\
\texttt{<eop>}\\
\imend \\
\asst\\ Thought: To complete the task, I need to locate two newspapers and place them in an armchair. I should start by searching the nearby furniture for any newspapers.\\Action: go to sofa 1\ \imend \\
\user\\ Observation: On the sofa 1, you see a box 3, a box 2, a newspaper 2, and a pillow 1.\ \imend \\
\asst\\ Thought: I've found one newspaper on sofa 1. I should pick it up to place it in the armchair later.\\Action: take newspaper 2 from sofa 1\ \imend \\
\user\\ Observation: You pick up the newspaper 2 from the sofa 1.\ \imend \\
\asst\\ Thought: With newspaper 2 in hand, I need to find the second newspaper. The dining table is another common place where newspapers might be found.\\Action: go to dining table 1\ \imend \\
\user\\ Observation: On the dining table 1, you see a box 1, a credit card 2, a credit card 1, a laptop 1, a newspaper 1, a plate 3, a plate 1, a remote control 1, a statue 3, a statue 1, a tissue box 3, and a tissue box 1.\ \imend \\
\asst\\ There's the second newspaper on the dining table. I should take it so I can place both newspapers in the armchair as instructed.\\Action: take newspaper 1 from dining table 1\ \imend \\
\user\\ Observation: You pick up the newspaper 1 from the dining table 1.\ \imend \\
\asst\\ Thought: With both newspapers now in my possession, I need to find an armchair to place them in.\\Action: go to armchair 1\ \imend \\
\user\\ Observation: On the armchair 1, you see a keychain 1.\ \imend \\
\asst\\ Thought: The armchair has a keychain on it, but there's enough space for the newspapers. I'll place the first newspaper here.\\Action: put newspaper 2 in/on armchair 1\ \imend \\
\user\\ Observation: You put the newspaper 2 in/on the armchair 1.\ \imend \\
\asst\\ Thought: With the first newspaper placed, I need to put the second newspaper in the armchair to complete the task.\\Action: put newspaper 1 in/on armchair 1\ \imend
\end{templatebox}

\end{document}